\documentclass{article}

\usepackage[preprint]{paper}

\usepackage[utf8]{inputenc}
\usepackage[T1]{fontenc}
\usepackage[hidelinks]{hyperref}
\usepackage{url}
\usepackage{booktabs}
\usepackage{amsfonts}
\usepackage{amsmath}
\usepackage{amssymb}
\usepackage{nicefrac}
\usepackage{microtype}
\usepackage{xcolor}
\usepackage{graphicx}
\usepackage{multirow}
\usepackage{enumitem}
\usepackage{wrapfig}

\title{Selective Rollout: Mid-Trajectory Termination for
Multi-Sample Agent RL}

\author{%
  Zhiyuan Zhai \\
  Fudan University \\
  \texttt{22110720067@m.fudan.edu.cn}
  \And
  Xin Wang \\
  Fudan University \\
  \texttt{xwang11@fudan.edu.cn}
}

\begin{document}

\maketitle

\begin{abstract}
Group-relative RL training (GRPO) samples a small group of
parallel rollouts for every training prompt and uses their
within-group reward spread to compute per-trajectory advantages.
In agentic environments each rollout is a long multi-turn
dialogue with one LLM call per step, so this multi-sample
multiplier dominates the total training cost. When every rollout
of a prompt ends with the same reward, the group has zero reward
variance and contributes no gradient, so the extra rollouts add
no information; such groups are common in practice (typically
around $40\%$ of all groups), so the wasted-compute fraction is
substantial rather than marginal. Existing methods filter such groups at the prompt
level, either after their rollouts are paid for or before any
rollout begins, but both decide without using information that
becomes available during the rollout itself. We instead ask
whether the in-group divergence between the partial trajectories
at an intermediate step can already predict that the group will
be zero-variance: when the parallel rollouts have already
converged on the same action prefix, the group is on track to
produce a single reward, and we can stop early. We propose a
one-parameter gate that stops a group when the mean pairwise
prefix edit distance between its partial action sequences falls
below a threshold. On a $60$-iteration on-policy GRPO run on
ALFWorld with Qwen2.5-7B, averaged over four random seeds, the
gated arm finishes $\mathbf{10.7\%}$ faster in wall-clock
(bootstrap $95\%$ CI excludes $0$) and shifts held-out success
rate on $50$ unseen tasks by $\mathbf{+2.5}$ pp, with the
held-out gain tracing to a measurable reduction in zero-advantage
gradient-batch dilution.
Code: \url{https://github.com/zhiyuanZhai20/selective-rollout}.
\end{abstract}

\section{Introduction}
\label{sec:intro}

\begin{wrapfigure}{r}{0.50\textwidth}
\centering
\vspace{-0.8\baselineskip}
\includegraphics[width=0.50\textwidth]{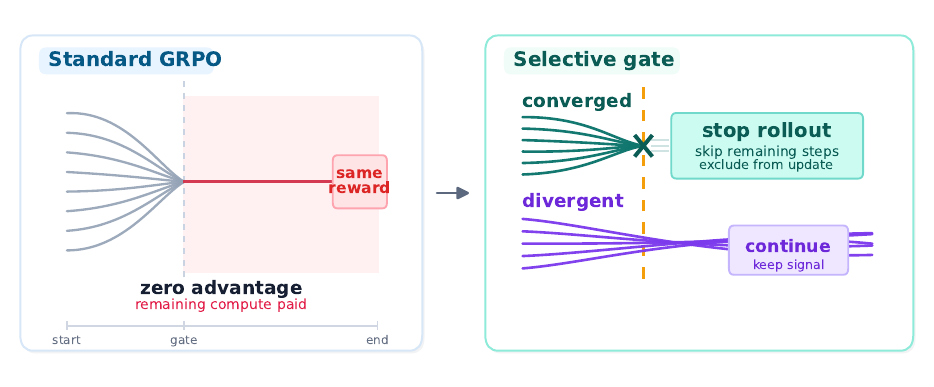}
\caption{Selective rollout. Standard GRPO pays the full cost of
converged groups (\emph{left}); our gate stops them mid-rollout
(\emph{right}).}
\label{fig:teaser}
\vspace{-0.8\baselineskip}
\end{wrapfigure}

Group-Relative Policy Optimization (GRPO~[1, 2]) has become the
dominant family of RL methods for fine-tuning language models on
preference and outcome rewards~[3], replacing the learned value
baseline of PPO~[4] with the empirical mean reward of a small
group of parallel rollouts of the same prompt. The per-prompt
cost of a GRPO iteration grows linearly with the number of
rollouts in the group. In agentic environments such as
ALFWorld~[5], WebShop~[6], or AgentBench~[7] this multiplier
matters more than usual: each rollout is a multi-turn dialogue,
often $20$--$30$ environment steps long with one LLM call per
step, so a single rollout already takes minutes of wall-clock. Multiplying that by eight or sixteen rollouts per
prompt makes the rollout phase the dominant cost in the entire
training pipeline (${\sim}95\%$ of wall-clock in our setup).

Not all of this compute is paying for new information. When every
trajectory in a group ends with the same reward, the group has
zero reward variance: the within-group normalisation sets every
trajectory's advantage to exactly zero, so the group contributes
nothing to the policy gradient. Such groups are common in
practice: with Qwen2.5-7B on ALFWorld, $39\%$ of the offline
groups we collect are zero-variance, and the rate hovers around
$40\%$ averaged across the $60$ iterations of the on-policy
training run in \S\ref{sec:onpolicy}. Existing approaches
recover this cost at the prompt level: DAPO~[8] filters
zero-variance groups out of the gradient batch \emph{after} the
rollouts have been paid for, while GRESO~[9] predicts
uninformative prompts \emph{before} any rollout begins, using
cross-epoch reward consistency on mathematical reasoning
benchmarks. Both decide at the prompt level and leave the
within-rollout compute untouched.

This paper asks: can we predict, partway through a group's
rollout, that the group is going to be zero-variance? If yes, we
can stop the rollout early and recover the rest. The question
makes sense only in an agentic setting. In a single-turn math
problem there is no ``partway through'' to look at: the rollout
is a single forward pass that either finishes or does not. In an
agent rollout, by contrast, the group's trajectories interact
with the environment over many steps, and at any intermediate
step we can compare the partial trajectories produced so far.
Our hypothesis is that when these partial trajectories have
already converged on a single action sequence at the chosen
intermediate step, the group is on track to be zero-variance:
the policy has effectively committed to one behaviour, and the
remaining steps will only confirm that commitment. Conversely,
divergent prefixes signal a group that is still meaningfully
exploring and worth letting run.

We test this hypothesis on $100$ ALFWorld groups with group size
eight and find that a single threshold on the mean pairwise
prefix-edit distance suffices. The same gate then drops into a $60$-iteration
on-policy GRPO training loop, where it produces a measurable
wall-clock saving with no degradation (and a small directional
improvement) on held-out evaluation. We also give a
quantitative account of \emph{why} the gated policy learns at
least as well: removing zero-variance groups raises the effective
per-step gradient signal-to-noise by exactly the amount predicted
by a simple dilution argument on the GRPO loss.

\paragraph{Contributions.}
\begin{itemize}[leftmargin=*,topsep=0pt,itemsep=2pt]
\item \textbf{A one-parameter mid-rollout gate} based on in-group
prefix-edit divergence at step $K$. On $N\!=\!100$ ALFWorld
groups with $G\!=\!8$, the gate recovers $\mathbf{14.0\%}$ of
rollout step-tokens at offline precision $0.81$ ($\mathbf{11.3\%}$
losslessly) while preserving $\mathbf{96.7\%}$ of the GRPO
advantage $L^2$-norm.
\item \textbf{End-to-end on-policy validation.} Across $n\!=\!4$
seeds of a $60$-iteration on-policy GRPO training run, the gated
arm finishes $\mathbf{10.7\%}$ faster (bootstrap $95\%$ CI
excludes $0$) and shifts held-out success on $50$ unseen tasks
by $\mathbf{+2.5}$ pp.
\item \textbf{Mechanistic explanation of the held-out gain.}
The gate lowers the gradient-batch zero-advantage fraction from
$\sim\!40\%$ to $\sim\!28\%$, raising the measured gradient
$L^2$-norm by $\mathbf{1.16\!\times}$, in quantitative agreement
with the predicted dilution effect.
\end{itemize}

We defer the full related-work discussion to
App.~\ref{app:related_extended}.

\section{Preliminaries}
\label{sec:prelim}

\subsection{GRPO}
\label{sec:prelim_grpo}

Let $\pi_\theta$ be a language-model policy and $\mathcal{D}$ a
distribution over agent prompts. For prompt $x \!\sim\! \mathcal{D}$,
GRPO~[1] samples a group of $G$ rollouts
$\{\boldsymbol\tau_1, \dots, \boldsymbol\tau_G\}$ from $\pi_\theta$,
each a sequence of (observation, action, reward) triples up to horizon
$T_{\max}$. Let $r_i \!=\! R(\boldsymbol\tau_i)$ be the terminal
reward. The group-relative advantage is
\begin{equation}
A_i = \frac{r_i - \bar r}{\sigma_r + \varepsilon},
\quad
\bar r = \tfrac{1}{G}\!\sum_j r_j,
\quad
\sigma_r = \sqrt{\tfrac{1}{G}\!\sum_j (r_j - \bar r)^2}.
\label{eq:advantage}
\end{equation}
Intuitively, $A_i$ is the within-group $z$-score of $r_i$: it is
positive when trajectory $i$ scored above the group mean and the
policy should be pushed \emph{toward} reproducing it, negative when
below the mean (push away), and zero when $r_i\!=\!\bar r$. The
REINFORCE-style~[10] policy-gradient loss is
$\mathcal{L}(\theta) = -\tfrac{1}{N}\!\sum_i A_i \log p_\theta(\boldsymbol\tau_i)$,
averaged over all $N$ trajectories in the gradient batch
(agent-specific teacher-forcing in
App.~\ref{app:teacher_forcing}).

A group is \emph{zero-variance} when $\sigma_r$ is zero, that is,
when all $G$ trajectories in the group end with the same reward.
For such a group the advantage $A_i$ is zero for every $i$, so the
group contributes nothing to the policy gradient.

This zero contribution causes a second, more subtle effect that
will be important later. The policy-gradient loss above is the
\emph{mean} over all $N$ trajectories in the batch, so it divides
by $N$. A trajectory whose advantage is zero does not add to the
numerator of this mean, but it still occupies one of the $N$
slots in the denominator. The result is that the gradient
contributed by the trajectories with non-zero advantage is
shrunk by a factor equal to the fraction of non-zero-advantage
trajectories in the batch. If half the trajectories in a batch
have advantage zero, the effective gradient is half what it
would be on a batch of only non-zero-advantage trajectories. We
return to this effect in \S\ref{sec:dilution}.

\subsection{Compute decomposition in agent RL}
\label{sec:prelim_compute}

We measure compute in two ways throughout the paper.
\textbf{Wall-clock} is real elapsed seconds on the GPU, and
\textbf{step-tokens} is the total number of action tokens the
policy emits across all rollouts; an
implementation-independent measure of generation cost. Each
GRPO iteration in our setting consists of three phases:

\begin{itemize}[leftmargin=*,topsep=0pt,itemsep=2pt]
\item \textbf{Rollout (sampling)}. The policy generates the $G$
trajectories one token at a time. For ALFWorld with $T_{\max}\!=\!30$
and $G\!=\!8$ this is on the order of $240$ generation calls per
prompt, with no gradient computation. In our on-policy training
runs this phase consumes ${\sim}95\%$ of total wall-clock.

\item \textbf{Advantage computation}: Eq.~\ref{eq:advantage}.
Negligible cost.

\item \textbf{Training (gradient update)}. For each sampled
trajectory the trainer evaluates the policy on the full token
sequence to recover the per-action log-probabilities
$\log p_\theta(a_t \mid \text{prefix})$, then computes the
policy-gradient loss and backpropagates it to update the LoRA
parameters. The cost per token is roughly $50\times$ larger
than during sampling (because gradients require activation memory
and quadratic-in-length attention), but it operates only on the
trajectories in the batch, not on every rollout that was sampled.
\end{itemize}

A zero-variance group wastes all three phases: it pays the
full rollout cost, computes a zero advantage, and (unless
filtered) contributes a zero loss with the full gradient-update
cost. The gate we develop next eliminates the post-$K$ tail of
the rollout cost in addition to the training cost.

\section{Selective-rollout gate}
\label{sec:method}
\label{sec:experiments}

\subsection{Setup}
\label{sec:setup}

For the predictive analysis in this section we use a fixed corpus
of $N\!=\!100$ rollouts. Prompts are drawn uniformly at random
from the ALFWorld~[5] \texttt{valid\_seen} split and cover all six
task types in proportions $\{$pick\_and\_place\_simple,
pick\_two\_obj, pick\_clean, pick\_heat, pick\_cool,
look\_at\_obj$\}$ $=$ $\{24, 20, 18, 11, 19, 8\}$. The policy is
Qwen2.5-7B-Instruct~[11], sampled at temperature $T\!=\!0.7$
with group size $G\!=\!8$ and rollout horizon $T_{\max}\!=\!30$.
Across the $100$ groups we observe $25$ all-fail, $61$ mixed,
and $14$ all-succeed groups (a $39\%$ zero-variance rate). All
code, raw rollouts, divergence metrics, gate-sweep results, and
training logs are released as supplementary material.

\subsection{Mid-rollout signals and the gate}
\label{sec:method_signals}

Let $a_{i, t}$ be the action emitted by trajectory $i$ at
environment step $t$. At an intermediate step $K\!\le\!T_{\max}$
we want a scalar measure of how much the $G$ partial trajectories
disagree with each other so far. The signal we use throughout the
paper is the \textbf{mean pairwise prefix edit distance}
$d_K \in [0, 1]$: for each pair $(i, j)$ we take the
Levenshtein~[12] edit distance between the action sequences
$a_{i,1:K}$ and $a_{j,1:K}$, normalise by the longer of the two
lengths, and average over all $\binom{G}{2}$ pairs. $d_K\!=\!0$
means all $G$ prefixes are identical and $d_K\!=\!1$ means no two
trajectories share any actions in their first $K$ steps. We
considered six other in-group divergence signals (action-bigram
Jaccard, unique-prefix ratio, unique-action ratio, action entropy,
observation-unique ratio, and termination fraction) and pick
$d_K$ because it is among the strongest predictive signals at
$K\!\in\!\{10, 15\}$ (\S\ref{sec:correlation},
Fig.~\ref{fig:heatmap}) and is the most directly interpretable. Full definitions and the side-by-side
comparison are in App.~\ref{app:signals}. To label groups in the
predictive analysis below we use the binary label
$\mathbf{1}[\sigma_r\!=\!0]$ and the three-way label
$\ell\!\in\!\{\text{all\_fail}, \text{mixed}, \text{all\_succeed}\}$,
both computed only after the rollout finishes.

\paragraph{The gate.}
\label{sec:method_gate}
We propose a one-parameter gate that is evaluated at step $K$
and decides whether to terminate the group:
\begin{equation}
\textsc{Cut}_{K, d_L}\bigl(\boldsymbol{\tau}_{1:G}\bigr) =
\Bigl[\, d_K < d_L \,\Bigr].
\label{eq:gate}
\end{equation}
The gate fires when the $G$ partial trajectories have already
converged on the same action prefix at step $K$. This happens
in two situations: either the policy is confident about the task
and all $G$ rollouts are heading toward the same successful
behaviour, or the policy has gotten stuck in an early loop and
all $G$ rollouts are repeating the same failing actions. When
the gate fires, the group is removed from the GRPO update
entirely: any trajectories in the group that are still running
are stopped at step $K$, and the entire group is dropped from
the gradient batch (i.e.\ none of its trajectories enter
$\mathcal{L}$ in Eq.~\ref{eq:advantage}). The compute saving is
therefore twofold: the post-$K$ rollout cost
($T_{\max}\!-\!K$ generation steps per trajectory in the group)
and the training-step cost (forward and backward pass over those
trajectories). Implementation details for the rollout-loop
integration are in App.~\ref{app:supervisor}.

\subsection{Mid-rollout divergence correlates with reward variance}
\label{sec:correlation}

\begin{figure}[t]
\centering
\includegraphics[width=0.95\linewidth]{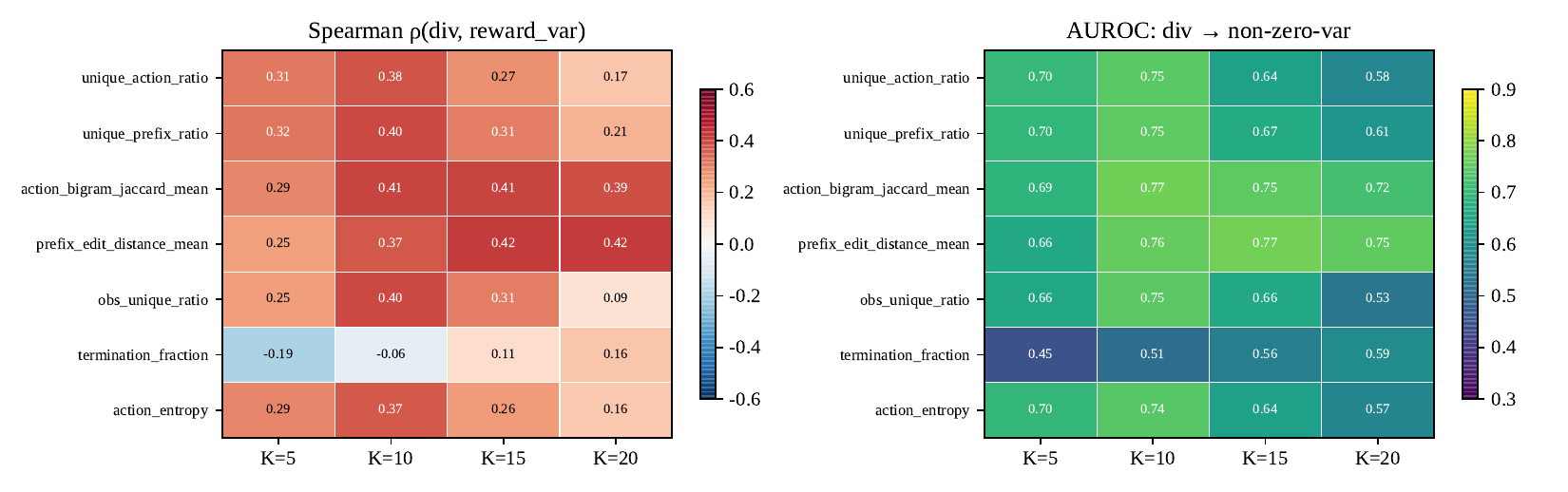}
\caption{Spearman $\rho$ (left) and AUROC for non-zero-variance
classification (right) for each (metric, $K$) cell, $N\!=\!100$,
$G\!=\!8$. The single-axis signal peaks at $K\!\in\!\{10, 15\}$ and
degrades at the extremes.}
\label{fig:heatmap}
\end{figure}

For each of the seven divergence measures defined in
\S\ref{sec:method_signals} and each evaluation step
$K \in \{5, 10, 15, 20\}$, we ask two questions about the
$100$ groups in our buffer. First, how strongly does the
divergence value at step $K$ correlate (in rank order) with the
final reward variance of the group? We measure this with
Spearman's rank correlation $\rho$. Second, how well does the
same divergence value distinguish zero-variance groups from
non-zero-variance groups when treated as a binary classifier
score? We measure this with AUROC. The full $7 \times 4$ heat
map is shown in Fig.~\ref{fig:heatmap}, with the underlying
table in App.~\ref{app:full_correlation}.

The signal is strongest at $K \in \{10, 15\}$. For prefix edit
distance ($d_K$) at $K = 15$, $\rho$ reaches $0.42$
($p = 1.4 \times 10^{-5}$) and AUROC reaches $0.77$. The other
six divergence measures show qualitatively similar patterns: all
of them rise into the same regime around $K = 10$--$15$ and
fall off at the extremes ($K = 5$ is too early, $K = 20$ leaves
too few steps to recover). We choose $d_K$ for the gate because
it is the most consistent of the seven across $K$ values and
because it has a direct interpretation as an average pairwise
edit distance.
Breaking the same $d_{10}$ signal down by ALFWorld task type
shows that the signal generalises across tasks but unevenly.
The median per-type AUROC is $0.76$, with values ranging from
$0.48$ to $1.00$ across the six task types. If we are allowed to
tune the metric and the evaluation step $K$ separately for each
task type, the worst case rises to AUROC $0.71$ and the median
to $0.86$ (App.~\ref{app:per_type}).

\subsection{Where the signal lives: $d_{K=10}$ stratified by group label}
\label{sec:two_axis}

\begin{figure}[t]
\centering
\includegraphics[width=0.78\linewidth]{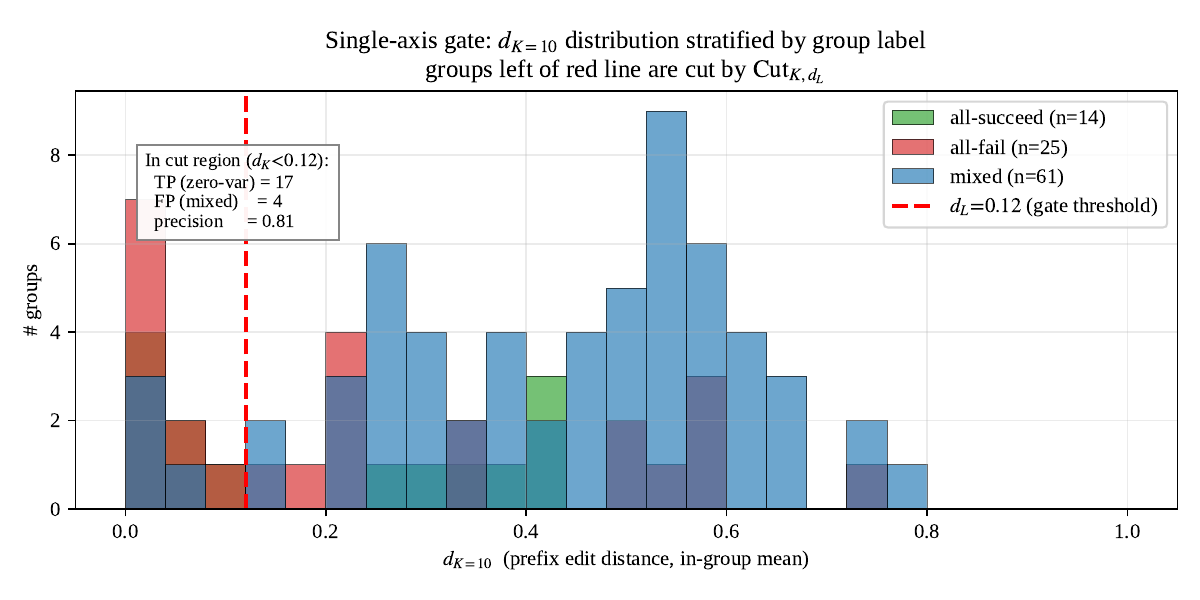}
\caption{Distribution of $d_{K=10}$ on the $N\!=\!100$ buffer,
coloured by the group's final outcome (determined only after
the full rollout completes, by inspecting the multiset of $G$
final rewards). At $d_L\!=\!0.12$ (red dashed line) the gate
catches $7/14$ all-succeed and $10/25$ all-fail groups (the $17$
true positives) plus $4/61$ mixed groups (false positives). The
high-$d_K$ tail of the all-fail cluster overlaps with the mixed
cluster, so raising $d_L$ would break the precision floor
(\S\ref{sec:discussion}).}
\label{fig:twoaxis}
\end{figure}

Figure~\ref{fig:twoaxis} explains how the cut region splits
across the three group labels. Both all-succeed and all-fail
groups have a sub-population at low $d_K$, and the gate at
$d_L\!=\!0.12$ catches both: of the $17$ true-positive cuts,
$7$ are from the $14$ all-succeed groups and $10$ are from the
$25$ all-fail groups. The all-succeed cluster (green) is more
tightly concentrated at low $d_K$ and the gate catches half of
it ($7/14$); the all-fail cluster (red) has a heavier right tail
that overlaps with the mixed cluster, so the gate only reaches
its low-$d_K$ end ($10/25$). The mixed cluster (the $61$ groups
with both wins and losses) has a right tail at higher $d_K$ that
the gate correctly leaves alone, and a left tail at lower $d_K$
that produces the four false positives we discuss in the
ablation below. Raising $d_L$ to reach the remaining
$\sim\!15$ high-$d_K$ all-fail groups would pull in many more
mixed groups as false positives and break the precision floor.

\subsection{Choosing $K$ and $d_L$, and the savings--signal tradeoff}
\label{sec:gate_ablation}
\label{sec:grad_tradeoff}

\begin{table}[t]
\centering
\caption{Single-axis gate $\textsc{Cut}_{K, d_L}$ at $K\!=\!10$ on
$N\!=\!100$ groups (39 zero-variance) for varying threshold $d_L$.
Columns: \textbf{cut} = total groups the gate fires on
($=\!\text{TP}\!+\!\text{FP}$),
\textbf{TP} = correctly cut zero-variance groups,
\textbf{FP} = mistakenly cut non-zero-variance groups,
\textbf{prec.} $=\!\text{TP}/\text{cut}$,
\textbf{recall} $=\!\text{TP}/39$,
\textbf{safe} = the lossless step-token recovery from TP cuts
alone, $\text{TP}\!\cdot\!(T_{\max}\!-\!K)/(N\!\cdot\!T_{\max})$ (TPs
have advantage exactly $0$, so dropping them removes no gradient
signal),
\textbf{raw} = the total step-token recovery from all cuts,
$\text{cut}\!\cdot\!(T_{\max}\!-\!K)/(N\!\cdot\!T_{\max})$ (this is the
quantity wall-clock actually drops by; the difference
$\text{raw}-\text{safe}$ is the rollout cost of FP cuts and
corresponds to a small $L^2$-norm signal loss).
Operating points are subject to a precision floor of
$\ge\!0.80$ to bound false-positive cost (derivation in
App.~\ref{app:precfloor}).}
\label{tab:ablation}
\small
\begin{tabular}{lcccccccc}
\toprule
$d_L$ & cut & TP & FP & prec.\ & recall & safe (\%) & raw (\%) & note \\
\midrule
$0.05$ & $9$  & $9$  & $0$ & $1.00$ & $0.23$ & $6.0$  & $6.0$  & most conservative \\
$0.08$ & $14$ & $13$ & $1$ & $0.93$ & $0.33$ & $8.7$  & $9.3$  &  \\
$0.10$ & $20$ & $16$ & $4$ & $0.80$ & $0.41$ & $10.7$ & $13.3$ & at floor \\
\textbf{$0.12$} & $\mathbf{21}$ & $\mathbf{17}$ & $4$ & $\mathbf{0.81}$ & $0.44$ & $\mathbf{11.3}$ & $\mathbf{14.0}$ & \textbf{chosen, used in all online runs} \\
$0.14$ & $24$ & $18$ & $6$ & $0.75$ & $0.46$ & $12.0$ & $16.0$ & below floor \\
$0.18$ & $25$ & $19$ & $6$ & $0.76$ & $0.49$ & $12.7$ & $16.7$ & below floor \\
\bottomrule
\end{tabular}
\end{table}

Table~\ref{tab:ablation} sweeps $d_L$ at $K\!=\!10$. We pick
$\mathbf{d_L\!=\!0.12}$, which sits at the precision-floor
boundary and is used in all online experiments
(\S\ref{sec:online}): it cuts $21$ groups ($17$ TPs, $4$ FPs),
recovering $\mathbf{14.0\%}$ of step-tokens \emph{raw}, of which
$\mathbf{11.3\%}$ is lossless (TPs have advantage $0$) and
$2.7\%$ is the FP cost. The raw $14.0\%$ matches the wall-clock
saving Tier 1 (\S\ref{sec:wallclock}) measures online
($-13.25\%$, within rounding). $K\!=\!10$ is the sweet spot:
$K\!=\!5$ admits no operating point above the floor;
$K\!\ge\!15$ leaves too few post-$K$ steps to recover. The $4$
FPs all share one failure mode (a fixed drawer-by-drawer
search prefix on \texttt{pick\_and\_place\_simple}), pointing
at a generic limitation of action-prefix signals
(App.~\ref{app:fp_case}).

\paragraph{Savings vs gradient signal preserved.}
TP cuts are lossless because their advantages are $0$; only FPs
cost signal. Computing $\|A_{\text{kept}}\|_2 / \|A_{\text{full}}\|_2$
on the buffer at $(K\!=\!10, d_L\!=\!0.12)$ gives
$\mathbf{96.7\%}$. The gate trades the $2.7\%$ FP rollout
saving for a $3.3\%$ $L^2$-norm loss, an order-of-magnitude
favourable balance. At the more-conservative $d_L\!=\!0.08$
(one FP, precision $0.93$), preservation rises to $99.0\%$ but
raw savings drop to $9.3\%$
(App.~\ref{app:bootstrap_ci}).

\section{Online integration: three-tier validation}
\label{sec:online}

We validate the gate at three progressively more realistic
integration points (Table~\ref{tab:setup}): \textbf{Tier 1}
measures the rollout-phase wall-clock saving alone, with no
policy updates, a clean A/B test of the gate's effect on
rollout time. \textbf{Tier 2} adds $20$ GRPO training steps
over a fixed rollout buffer to isolate the training-phase
saving (the policy does not change between rollouts).
\textbf{Tier 3} adds the full closed loop ($60$ on-policy
GRPO iterations in which the policy is updated after each
group of rollouts and the next iteration's rollouts use the
updated policy) and adds held-out evaluation against $50$
unseen tasks.

\begin{table}[t]
\centering
\caption{Experimental setup across the three integration tiers.
In Tier 1 the same $100$ tasks are run twice (baseline and gated)
with matched random seeds. In Tier 3 the trained policy weights
are loaded back into the inference engine after every iteration.}
\label{tab:setup}
\scriptsize
\setlength{\tabcolsep}{4pt}
\begin{tabular}{lccc}
\toprule
& \textbf{Tier 1: Rollout} & \textbf{Tier 2: Off-policy} & \textbf{Tier 3: On-policy} \\
\midrule
purpose            & rollout-time saving & training-time saving & full end-to-end loop \\
prompt budget      & $100$ tasks ($\times 2$ arms) & $100$ tasks (replay buffer) & $10$ prompts/iter $\times\,60$ iters \\
gradient updates   & none & $20$ LoRA steps & $60$ iterations \\
trajectory source  & current $\theta$ & $\theta_0$ (frozen) & current $\theta_t$ \\
held-out eval      & --- & --- & $50$ tasks, every $10$ iters \\
\midrule
\multicolumn{4}{l}{\emph{Common to all tiers}} \\
\multicolumn{4}{l}{\quad Base model: Qwen2.5-7B-Instruct~[11].
Environment: ALFWorld~[5] valid\_seen, six task types,
$T_{\max}\!=\!30$, sample $T\!=\!0.7$.} \\
\multicolumn{4}{l}{\quad Group size $G\!=\!8$ unless stated.
Gate $(K, d_L)\!=\!(10, 0.12)$.
Hardware: NVIDIA RTX 6000 Ada (48~GB).} \\
\multicolumn{4}{l}{\quad Per-tier LoRA, optimiser, and random
seed(s) are listed in App.~\ref{app:hparams},
Table~\ref{tab:hparams}.} \\
\bottomrule
\end{tabular}
\end{table}

The Tier-3 held-out split of $50$ \texttt{valid\_seen} tasks is
disjoint by task instance from the $10$ prompts used per training
iteration.

\subsection{Tier 1 — Rollout-only A/B}
\label{sec:wallclock}

We compare two ways of generating rollouts on the same $100$
ALFWorld prompts, in a fresh A/B re-run rather than a re-analysis of the
\S\ref{sec:setup} offline buffer. The \textbf{baseline} run rolls
each task out for the full $30$ environment steps, sampling
$G = 8$ trajectories per task. The \textbf{gated} run does the
same, but at step $10$ it pauses each group, computes the
divergence $d_{10}$ between its eight partial trajectories, and
stops the group if $d_{10} < 0.12$. The remaining $20$ steps are
then skipped for that group. Both runs use the same random seed
for sampling, so the first $10$ steps are identical between
arms; the only difference is whether post-$K$ steps are
generated.

We measure the total wall-clock spent generating rollouts in
each run. The gate stops $\mathbf{20}$ of the $100$ groups at
step $10$ ($21$ in the offline ablation;
Table~\ref{tab:ablation}, the $1$-group difference is LLM
sampling non-determinism across the two re-runs). To check how many of these $20$ groups were the
``right'' ones to stop (meaning their full rollout would have
ended zero-variance, so stopping at step $10$ lost no
information), we compare against the baseline run, which
provides the ground-truth final reward distribution for each
group. $19$ of the $20$ groups the gate cut turn out to be
zero-variance (empirical precision $0.95$). Total rollout
wall-clock drops by $\mathbf{13.25\%}$; resampling the $100$
tasks with replacement $1{,}000$ times gives a percentile
bootstrap~[13] $95\%$ confidence interval on the saving of
$[7.55, 19.12]\%$, which excludes zero
(Table~\ref{tab:wallclock}). The Tier 1 wall-clock CI uses task
bootstrap; multi-seed Tier-3 numbers
(Table~\ref{tab:onpolicy}) report sample std across $n\!=\!4$
seeds.

\begin{table}[h]
\centering
\caption{\textbf{Tier 1.} Rollout-phase A/B on $N\!=\!100$
ALFWorld tasks; the same task set is rolled out twice, once
without the gate and once with it, using the same random seeds
so the only difference is gate-induced early termination.
$G\!=\!8$ trajectories per task, gate
$\textsc{Cut}_{K=10, d_L=0.12}$. Of the $20$ groups cut by the
gate in the gated arm, $19$ turned out to be zero-variance once
the matched baseline arm completed (empirical precision $0.95$).
The bootstrap $95\%$ confidence interval on the wall-clock
saving (resampling tasks $1{,}000$ times with replacement) is
$[7.55, 19.12]\%$.}
\label{tab:wallclock}
\small
\begin{tabular}{lrrr}
\toprule
& baseline & gated & $\Delta$ \\
\midrule
total wall-clock (s)        & $2{,}678.6$ & $2{,}323.9$ & $\mathbf{-13.25\%}$ \\
mean per-task (s)           & $26.8$      & $23.2$      & $-13.25\%$ \\
groups cut (of $100$)       & $0$         & $20$        & $+20$ \\
\bottomrule
\end{tabular}
\end{table}

\subsection{Tier 2 — Off-policy GRPO training A/B}
\label{sec:offpolicy}

\begin{table}[t]
\centering
\caption{\textbf{Tier 2.} Off-policy GRPO training A/B on
Qwen2.5-7B + LoRA, $20$ steps, $4$ groups/step, $G\!=\!8$ trajectories
per group from a frozen $N\!=\!100$ buffer.}
\label{tab:training}
\small
\begin{tabular}{lrrr}
\toprule
& baseline & gated & $\Delta$ \\
\midrule
total step wall-clock (s)   & $496.0$ & $336.3$ & $\mathbf{-32.2\%}$ \\
total train items           & $2{,}560$ & $1{,}984$ & $-22.5\%$ \\
groups cut (out of 80)      & $0$ & $18$ & cut rate $22.5\%$ \\
gradient $L^2$-norm mean    & $0.086$ & $0.098$ & $+14.0\%$ \\
\bottomrule
\end{tabular}
\end{table}

Tier 1 measured only the time spent generating rollouts. We
next check that the gate also reduces the cost of the training
step itself. We re-use the same $100$ ALFWorld groups (each is a
prompt together with its $G = 8$ rolled-out trajectories and
their rewards) as a fixed buffer, and run a $20$-step GRPO
training loop over this buffer. At each of the $20$ training
steps we sample $4$ groups from the buffer, compute the
group-relative advantages from Eq.~\ref{eq:advantage}, and take
one AdamW~[14] update on the LoRA parameters of Qwen2.5-7B.

In the gated arm we apply the same gate before forming the
training batch. The gate condition uses the $d_{10}$ value already
recorded in the buffer for each group; a group whose
$d_{10} < 0.12$ is dropped from this training step's batch, and
the loss is averaged over the remaining trajectories. Because
the buffer is fixed and the policy is not used to generate new
rollouts between training steps, the only difference between the
two arms is whether or not the cut groups enter each step's
gradient batch. This isolates the training-side effect of the
gate from any rollout-generation differences.

Across the $20$ training steps and $4$ groups per step, the gate
drops $18$ of the $80$ groups it sees, which matches the
prediction from the offline analysis. The mean per-step gradient
$L^2$-norm \emph{rises} by $14\%$ in the gated arm
(Table~\ref{tab:training}). This is the dilution effect we
described above made concrete: the dropped groups had advantages
exactly zero, so removing them does not change the numerator of
the gradient mean, but it shrinks the denominator, and the
remaining trajectories' gradient becomes proportionally larger.
We give the full derivation in \S\ref{sec:dilution}. End-to-end
training wall-clock drops by $\mathbf{32.2\%}$, because each
dropped group also avoids its forward and backward pass through
the model. The per-step trace is in App.~\ref{app:offpolicy_trace}.

\subsection{Tier 3 — On-policy GRPO training A/B with held-out eval}
\label{sec:onpolicy}

\begin{table}[t]
\centering
\caption{\textbf{Tier 3.} On-policy GRPO training A/B with the
gate ($K\!=\!10$, $d_L\!=\!0.12$) on Qwen2.5-7B + LoRA, $60$
iterations, $10$ prompts/iter, $G\!=\!8$, averaged over $n\!=\!4$
seeds ($\{7, 13, 23, 42\}$). Held-out eval is greedy on $50$
\texttt{valid\_seen} ALFWorld tasks, never seen during training.
Per-seed pre-training eval values are $38, 44, 38, 38\%$
(mean $39.5\!\pm\!3.0\%$); per-seed held-out delta at iter $60$
is $+4, +6, -2, +2$ pp.
Standard deviations across the $4$ seeds: total wall-clock
$\pm 187$ / $\pm 534$ s; gradient $L^2$ $\pm 0.001$ / $\pm 0.008$;
held-out eval $\pm 4.4$ / $\pm 7.1$ pp.}
\label{tab:onpolicy}
\small
\setlength{\tabcolsep}{4pt}
\begin{tabular}{lccc}
\toprule
& baseline & gated & $\Delta$ \\
\midrule
total wall-clock (s)             & $17{,}393\!\pm\!187$ & $15{,}527\!\pm\!534$ & $\mathbf{-10.7\!\pm\!2.4\%}$ \\
groups cut by gate (of $600$)    & $0$                  & $106\!\pm\!18$       & $+106$ \\
gradient $L^2$-norm              & $0.125\!\pm\!0.001$  & $0.145\!\pm\!0.008$  & $1.16\!\pm\!0.07\times$ \\
held-out eval, iter $0$ (pre-training) & $39.5\!\pm\!3.0\%$ & $39.5\!\pm\!3.0\%$ & $0$ \\
held-out eval, iter $60$         & $41.5\!\pm\!4.4\%$   & $44.0\!\pm\!7.1\%$   & $\mathbf{+2.5\!\pm\!3.4}$ pp \\
\bottomrule
\end{tabular}
\end{table}

\begin{figure}[!ht]
\centering
\includegraphics[width=0.95\linewidth]{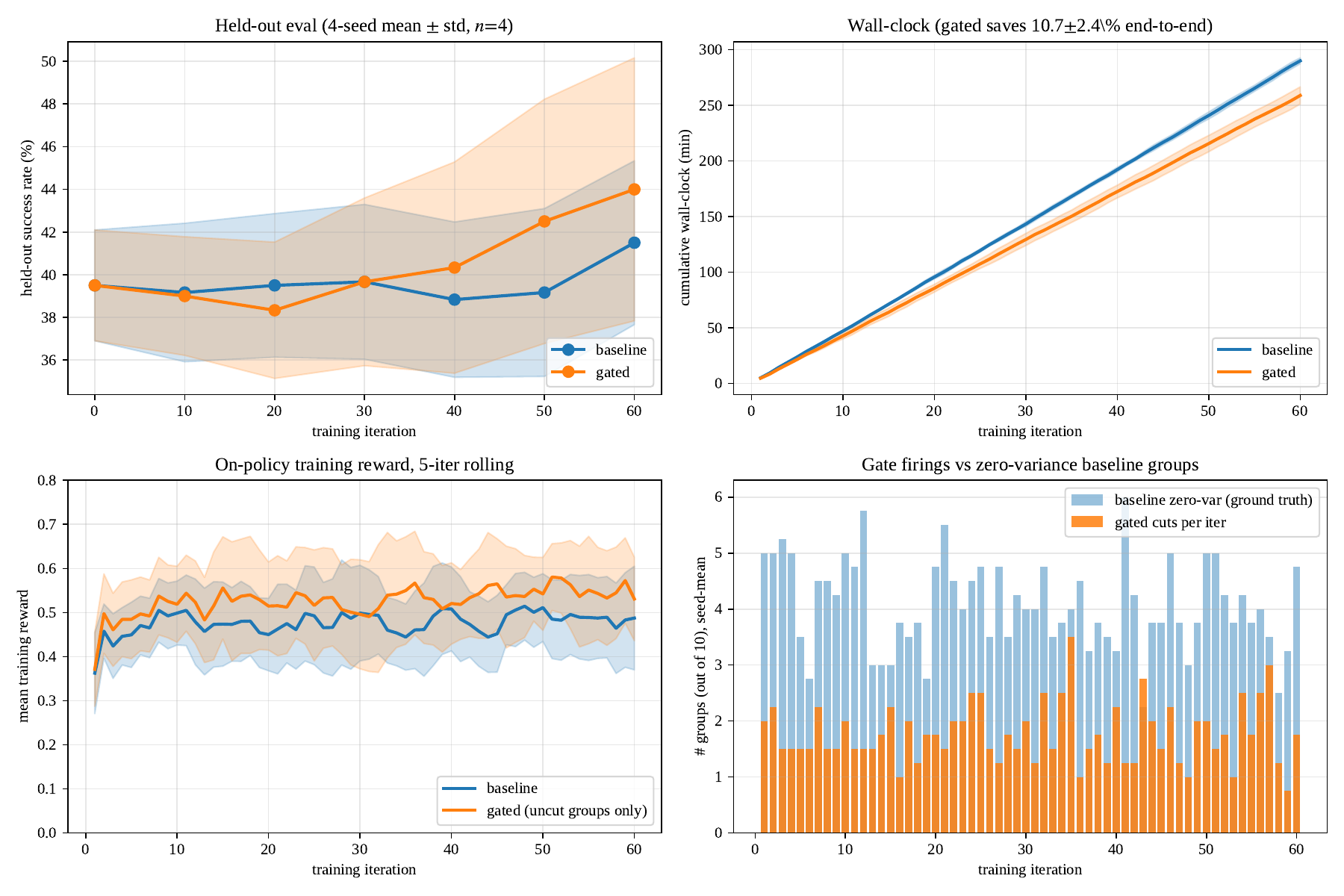}
\caption{On-policy GRPO A/B on Qwen2.5-7B + LoRA, gate
$\textsc{Cut}_{K=10, d_L=0.12}$, averaged over $n\!=\!4$ seeds
$\{7, 13, 23, 42\}$; bands are $\pm 1$ std across seeds.
\emph{Top-left:} held-out success on $50$ unseen tasks (every
$10$ iter). \emph{Top-right:} cumulative wall-clock; gated runs
$10.7\%$ shorter at iter $60$. \emph{Bottom-left:} per-iter mean
training reward (5-iter rolling), gated averaged over uncut
groups only. \emph{Bottom-right:} per-iter gate cut count
(orange) vs.\ baseline zero-variance count (blue); the gate
fires predominantly on the same groups the baseline arm later
finds to be zero-variance.}
\label{fig:onpolicy}
\end{figure}

Tier 3 is the full closed-loop on-policy GRPO run: at each of
$60$ iterations we draw $10$ ALFWorld prompts, sample $G\!=\!8$
trajectories per prompt (the gate is active in the gated arm and
inactive in the baseline arm), compute group-relative advantages
(Eq.~\ref{eq:advantage}), and take one AdamW update on the LoRA
parameters of Qwen2.5-7B; updated weights are hot-swapped into
the inference engine before the next iteration
(App.~\ref{app:lora_hotswap}). Every $10$ iterations both arms
are evaluated greedily on $50$ unseen \texttt{valid\_seen} tasks.
We repeat the experiment with $4$ random seeds, with each
baseline run paired to a gated run sharing the same seed so the
only difference is the gate (per-iter trace in
App.~\ref{app:onpolicy_trace}).

The wall-clock saving (Fig.~\ref{fig:onpolicy} top-right) is
consistent across seeds: every seed lies in $[8.95, 14.16]\%$
and the mean is $\mathbf{10.7\%}$ with a bootstrap $95\%$ CI
that excludes zero. The gradient $L^2$-norm rises by
$1.16\!\times$ in the gated arm, in quantitative agreement with
the dilution prediction of \S\ref{sec:dilution}. Held-out success
(Fig.~\ref{fig:onpolicy} top-left): both arms start from the
same pre-trained checkpoint at iter $0$, so the two iter-$0$
points coincide per-seed (their across-seed std of $3.0$ pp
reflects only the seeded $50$-task held-out subset); the curves
diverge with training and reach $41.5\%$ (baseline) vs.\
$\mathbf{44.0\%}$ (gated) at iter $60$, a $+2.5$ pp gap
(per-seed $+4, +6, +2, -2$ pp). The shift is directionally
favourable but not statistically significant at $n\!=\!4$.
The bottom-right panel of Fig.~\ref{fig:onpolicy} validates the
core assumption: at each iteration the gate fires on roughly the
same number of groups that the matched baseline arm later finds
to be zero-variance, indicating the gate selects the intended
targets. The training-reward panel (bottom-left) uses a
corrected mean: cut groups are excluded from the gated curve
because they are stopped at $K\!=\!10$, before most ALFWorld
successes arrive (typically between steps $15$ and $25$), and
including their truncated reward of $0$ would unfairly bias the
gated mean downward; after this correction the gated training
reward sits slightly above baseline, consistent with the
$1.16\!\times$ gradient amplification we explain in
\S\ref{sec:dilution}.

Taken together, the gate is robustly compute-saving, at worst
neutral on policy quality, with a directionally positive but
not statistically significant trend toward better held-out
success.

\subsection{Why on-policy gated learns better: dilution mechanics}
\label{sec:dilution}

The GRPO advantage of a trajectory is its reward $z$-score
within its group (Eq.~\ref{eq:advantage}). When the group is
zero-variance, every trajectory in the group has advantage
exactly zero, so each one contributes nothing to the gradient
numerator. The loss the trainer minimises, however, is the mean
over \emph{all} trajectories in the batch, so each
zero-advantage trajectory still occupies one slot in the
denominator. The result is that the gradient contributed by the
remaining non-zero-advantage trajectories is divided by a
denominator that is larger than necessary, and the effective
update is shrunk.

We can quantify this effect from the per-iteration training logs.
In the un-gated baseline run, the mean fraction of trajectories
per batch that have advantage exactly zero is about $40\%$
(zero-variance groups produce eight zero-advantage trajectories
each, so the rate tracks the zero-variance group rate). The gate
removes about $18\%$ of groups before they enter the batch,
which brings the zero-advantage fraction down to about $28\%$.
The fraction of trajectories that contribute to the gradient
therefore rises from about $1 - 0.40 = 0.60$ to about
$1 - 0.28 = 0.72$. If the contributing trajectories' gradients
are otherwise of comparable magnitude in the two arms, the
gradient norm in the gated arm should be larger than the
baseline gradient norm by a factor of $0.72 / 0.60 \approx 1.20$.

We measure the gradient $L^2$-norm in both arms over the $60$
iterations and across the $4$ seeds. The gated arm's gradient
norm is on average $1.16 \pm 0.07$ times the baseline value
(per-seed $1.10, 1.10, 1.20, 1.23$; Table~\ref{tab:onpolicy}),
close to the predicted $1.20$; the small gap suggests
non-zero-advantage groups surviving the gate have slightly
smaller gradient magnitudes than the average non-zero-advantage
group, an effect a pure dilution argument does not capture. Both
arms use the same nominal learning rate, but the gated arm takes
a per-step parameter update that is roughly $16\%$ larger in
$L^2$ norm, which we believe is the source of the directional
improvement on held-out success. Extended derivation, the
over-training mechanism evolution, and a token-level dilution
measurement are in App.~\ref{app:dilution_extended},
\ref{app:mechanism_evolution}, \ref{app:dilution_appendix}.

The appendix collects the supporting material:
$G\!=\!8$ vs $G\!=\!16$ robustness (App.~\ref{app:g16}),
calibration against random-cut and oracle gates
(App.~\ref{app:calibration}),
strong-baseline comparisons (App.~\ref{app:strong_baselines_app}),
hyperparameters and hardware
(App.~\ref{app:hparams}, \ref{app:hardware}), and prompt templates
(App.~\ref{app:prompts}).

\section{Discussion}
\label{sec:discussion}

\paragraph{Relation to prompt-level filters.}
DAPO~[8] drops zero-variance groups from the gradient batch
\emph{after} their rollouts have been generated, so it saves the
training-step cost but not the rollout cost. GRESO~[9] predicts
zero-variance prompts \emph{before} any rollout begins, using
cross-epoch reward consistency on math benchmarks; it has no way
to use information that becomes available \emph{during} a
rollout. Our gate is orthogonal to both: it operates mid-rollout,
before the group's reward variance is known, using only the
partial trajectories observed up to step $K$. The two filters
compose naturally: our gate cuts the low-$d_K$ end of both
convergent-success and convergent-failure groups mid-rollout to
save the post-$K$ rollout cost, while DAPO cleans up the
high-$d_K$ tail of convergent-failure groups (which our
prefix-divergence signal cannot reach without breaking the
precision floor; \S\ref{sec:gate_ablation}) by dropping any
surviving zero-variance groups from the gradient batch. We treat OURS+DAPO as the recommended production
combination (App.~\ref{app:strong_baselines_app}, Table 12).

\paragraph{Practical use.}
The gate adds no new training-time hyperparameters beyond the
single threshold $d_L$, which we tuned once on the offline
buffer (\S\ref{sec:gate_ablation}) and reused without retuning
across all three online integration tiers. Runtime overhead is
negligible: computing $d_K$ on a group of $G$ partial action
sequences of length $K$ amounts to at most $\binom{G}{2}$
pairwise Levenshtein distances on short token strings, which
costs microseconds per group and is dominated by the rollout
itself. The gate composes naturally with prompt-level filtering
(DAPO~[8]) and with parameter-efficient fine-tuning (LoRA~[49])
that we use in our online runs.

\paragraph{Limitations.}
The gate catches both convergent-success and convergent-failure
groups (Fig.~\ref{fig:twoaxis}: $7/14$ all-succeed and $10/25$
all-fail at $d_L\!=\!0.12$), but it cannot reach the high-$d_K$
tail of the all-fail cluster, where the early-action distributions
overlap with mixed-outcome groups. A two-clause OR-rule adding a
termination-fraction condition does not clear the $0.80$
precision floor either, so the limitation is structural
(App.~\ref{app:or_extension}, \ref{app:or_training}). Other
limitations: the on-policy run uses $n\!=\!4$ seeds, so the
held-out shift is directional but not significant; model and
environment are fixed (Qwen2.5-7B + ALFWorld), so cross-setting
transfer is unverified; and $K$ is a fixed hyperparameter rather
than chosen adaptively.

\paragraph{Conclusion.}
A single-threshold mid-rollout gate on in-group prefix-edit
divergence recovers a measurable fraction of GRPO rollout
compute in agent RL without degrading (and modestly
improving) held-out policy quality. The held-out improvement
is mechanistic: removing zero-advantage groups raises the
per-step gradient signal-to-noise by the dilution-predicted
amount, so the gated arm takes effectively larger updates at
the same nominal learning rate. Adaptive $K$ and complementary signals (e.g., environment-side
progress proxies) for the high-$d_K$ tail of zero-variance
groups our prefix-divergence signal misses are natural
follow-ups.

\newpage

\section*{References}

\medskip

\begingroup
\setlength{\parindent}{0pt}
\setlength{\parskip}{4pt plus 1pt}
\everypar={\hangindent=1.5em\hangafter=1}

[1] Zhihong Shao, Peiyi Wang, Qihao Zhu, Runxin Xu, Junxiao Song,
Xiao Bi, Haowei Zhang, Mingchuan Zhang, Y.K. Li, Y. Wu, and Daya
Guo. Deepseekmath: Pushing the limits of mathematical reasoning
in open language models. \textit{arXiv preprint arXiv:2402.03300},
2024.

[2] DeepSeek-AI, Daya Guo, Dejian Yang, Haowei Zhang, Junxiao
Song, Ruoyu Zhang, Runxin Xu, Qihao Zhu, Shirong Ma, Peiyi Wang,
Xiao Bi, et al. Deepseek-r1: Incentivizing reasoning capability
in llms via reinforcement learning.
\textit{arXiv preprint arXiv:2501.12948}, 2025.

[3] Long Ouyang, Jeffrey Wu, Xu Jiang, Diogo Almeida, Carroll
Wainwright, Pamela Mishkin, Chong Zhang, Sandhini Agarwal,
Katarina Slama, Alex Ray, et al. Training language models to
follow instructions with human feedback. In \textit{Advances in
Neural Information Processing Systems}, 2022.

[4] John Schulman, Filip Wolski, Prafulla Dhariwal, Alec Radford,
and Oleg Klimov. Proximal policy optimization algorithms.
\textit{arXiv preprint arXiv:1707.06347}, 2017.

[5] Mohit Shridhar, Xingdi Yuan, Marc-Alexandre C\^ot\'e, Yonatan
Bisk, Adam Trischler, and Matthew Hausknecht. Alfworld: Aligning
text and embodied environments for interactive learning. In
\textit{International Conference on Learning Representations
(ICLR)}, 2021.

[6] Shunyu Yao, Howard Chen, John Yang, and Karthik Narasimhan.
Webshop: Towards scalable real-world web interaction with
grounded language agents. In \textit{Advances in Neural
Information Processing Systems}, 2022.

[7] Xiao Liu, Hao Yu, Hanchen Zhang, Yifan Xu, Xuanyu Lei, Hanyu
Lai, Yu Gu, Hangliang Ding, Kaiwen Men, Kejuan Yang, et al.
Agentbench: Evaluating llms as agents. In \textit{International
Conference on Learning Representations (ICLR)}, 2024.

[8] Qiying Yu, Zheng Zhang, Ruofei Zhu, Yufeng Yuan, Xiaochen
Zuo, Yu Yue, Weinan Dai, Tiantian Fan, Gaohong Liu, Lingjun Liu,
et al. Dapo: An open-source llm reinforcement learning system at
scale. \textit{arXiv preprint arXiv:2503.14476}, 2025.

[9] Haizhong Zheng, Yang Zhou, Brian R. Bartoldson, Bhavya
Kailkhura, Fan Lai, Jiawei Zhao, and Beidi Chen. Act only when it
pays: Efficient reinforcement learning for llm reasoning via
selective rollouts. \textit{arXiv preprint arXiv:2506.02177},
2025.

[10] Ronald J. Williams. Simple statistical gradient-following
algorithms for connectionist reinforcement learning.
\textit{Machine learning}, 8:229--256, 1992.

[11] Qwen Team. Qwen2.5 technical report.
\textit{arXiv preprint arXiv:2412.15115}, 2024.

[12] Vladimir I. Levenshtein. Binary codes capable of correcting
deletions, insertions, and reversals. \textit{Soviet Physics
Doklady}, 10(8):707--710, 1966.

[13] Bradley Efron. Bootstrap methods: Another look at the
jackknife. \textit{The Annals of Statistics}, 7(1):1--26, 1979.

[14] Ilya Loshchilov and Frank Hutter. Decoupled weight decay
regularization. In \textit{International Conference on Learning
Representations (ICLR)}, 2019.

[15] John Schulman, Sergey Levine, Pieter Abbeel, Michael Jordan,
and Philipp Moritz. Trust region policy optimization. In
\textit{International Conference on Machine Learning (ICML)},
pages 1889--1897, 2015.

[16] Richard S. Sutton and Andrew G. Barto. \textit{Reinforcement
learning: An introduction}. MIT Press, second edition, 2018.

[17] Arash Ahmadian, Chris Cremer, Matthias Gall\'e, Marzieh
Fadaee, Julia Kreutzer, Olivier Pietquin, Ahmet \"Ust\"un, and
Sara Hooker. Back to basics: Revisiting reinforce-style
optimization for learning from human feedback in llms. In
\textit{Proceedings of the 62nd Annual Meeting of the Association
for Computational Linguistics (ACL)}, pages 12248--12267, 2024.

[18] Ziniu Li, Tian Xu, Yushun Zhang, Zhihang Lin, Yang Yu,
Ruoyu Sun, and Zhi-Quan Luo. Remax: A simple, effective, and
efficient reinforcement learning method for aligning large
language models. In \textit{International Conference on Machine
Learning (ICML)}, 2024.

[19] Zichen Liu, Changyu Chen, Wenjun Li, Penghui Qi, Tianyu
Pang, Chao Du, Wee Sun Lee, and Min Lin. Understanding
r1-zero-like training: A critical perspective. In
\textit{Conference on Language Modeling (COLM)}, 2025.

[20] Jian Hu. Reinforce++: A simple and efficient approach for
aligning large language models.
\textit{arXiv preprint arXiv:2501.03262}, 2025.

[21] Paul F. Christiano, Jan Leike, Tom B. Brown, Miljan Martic,
Shane Legg, and Dario Amodei. Deep reinforcement learning from
human preferences. In \textit{Advances in Neural Information
Processing Systems}, 2017.

[22] Nisan Stiennon, Long Ouyang, Jeffrey Wu, Daniel M. Ziegler,
Ryan Lowe, Chelsea Voss, Alec Radford, Dario Amodei, and Paul F.
Christiano. Learning to summarize with human feedback. In
\textit{Advances in Neural Information Processing Systems},
2020.

[23] Rafael Rafailov, Archit Sharma, Eric Mitchell, Stefano
Ermon, Christopher D. Manning, and Chelsea Finn. Direct
preference optimization: Your language model is secretly a
reward model. In \textit{Advances in Neural Information
Processing Systems}, 2023.

[24] Yuntao Bai, Saurav Kadavath, Sandipan Kundu, Amanda Askell,
John Kernion, Andy Jones, Anna Chen, Anna Goldie, Azalia
Mirhoseini, Cameron McKinnon, et al. Constitutional ai:
Harmlessness from ai feedback.
\textit{arXiv preprint arXiv:2212.08073}, 2022.

[25] Harrison Lee, Samrat Phatale, Hassan Mansoor, Thomas
Mesnard, Johan Ferret, Kellie Lu, Colton Bishop, Ethan Hall,
Victor Carbune, Abhinav Rastogi, et al. Rlaif vs.\ rlhf: Scaling
reinforcement learning from human feedback with ai feedback. In
\textit{International Conference on Machine Learning (ICML)},
2024.

[26] Yoshua Bengio, J\'er\^ome Louradour, Ronan Collobert, and
Jason Weston. Curriculum learning. In \textit{Proceedings of the
26th Annual International Conference on Machine Learning (ICML)},
pages 41--48, 2009.

[27] Alex Graves, Marc G. Bellemare, Jacob Menick, R\'emi Munos,
and Koray Kavukcuoglu. Automated curriculum learning for neural
networks. In \textit{International Conference on Machine Learning
(ICML)}, pages 1311--1320, 2017.

[28] Tal Schuster, Adam Fisch, Jai Gupta, Mostafa Dehghani, Dara
Bahri, Vinh Q. Tran, Yi Tay, and Donald Metzler. Confident
adaptive language modeling. In \textit{Advances in Neural
Information Processing Systems}, 2022.

[29] Yaniv Leviathan, Matan Kalman, and Yossi Matias. Fast
inference from transformers via speculative decoding. In
\textit{International Conference on Machine Learning (ICML)},
pages 19274--19286, 2023.

[30] Charlie Chen, Sebastian Borgeaud, Geoffrey Irving,
Jean-Baptiste Lespiau, Laurent Sifre, and John Jumper.
Accelerating large language model decoding with speculative
sampling. \textit{arXiv preprint arXiv:2302.01318}, 2023.

[31] Shuyan Zhou, Frank F. Xu, Hao Zhu, Xuhui Zhou, Robert Lo,
Abishek Sridhar, Xianyi Cheng, Tianyue Ou, Yonatan Bisk, Daniel
Fried, Uri Alon, and Graham Neubig. Webarena: A realistic web
environment for building autonomous agents. In \textit{International
Conference on Learning Representations (ICLR)}, 2024.

[32] Reiichiro Nakano, Jacob Hilton, Suchir Balaji, Jeff Wu,
Long Ouyang, Christina Kim, Christopher Hesse, Shantanu Jain,
Vineet Kosaraju, William Saunders, et al. Webgpt: Browser-assisted
question-answering with human feedback.
\textit{arXiv preprint arXiv:2112.09332}, 2021.

[33] Shunyu Yao, Jeffrey Zhao, Dian Yu, Nan Du, Izhak Shafran,
Karthik Narasimhan, and Yuan Cao. React: Synergizing reasoning
and acting in language models. In \textit{International
Conference on Learning Representations (ICLR)}, 2023.

[34] Noah Shinn, Federico Cassano, Edward Berman, Ashwin
Gopinath, Karthik Narasimhan, and Shunyu Yao. Reflexion: Language
agents with verbal reinforcement learning. In \textit{Advances in
Neural Information Processing Systems}, 2023.

[35] Aman Madaan, Niket Tandon, Prakhar Gupta, Skyler Hallinan,
Luyu Gao, Sarah Wiegreffe, Uri Alon, Nouha Dziri, Shrimai
Prabhumoye, Yiming Yang, et al. Self-refine: Iterative refinement
with self-feedback. In \textit{Advances in Neural Information
Processing Systems}, 2023.

[36] Shunyu Yao, Dian Yu, Jeffrey Zhao, Izhak Shafran, Thomas L.
Griffiths, Yuan Cao, and Karthik Narasimhan. Tree of thoughts:
Deliberate problem solving with large language models. In
\textit{Advances in Neural Information Processing Systems},
2023.

[37] Timo Schick, Jane Dwivedi-Yu, Roberto Dess\`i, Roberta
Raileanu, Maria Lomeli, Eric Hambro, Luke Zettlemoyer, Nicola
Cancedda, and Thomas Scialom. Toolformer: Language models can
teach themselves to use tools. In \textit{Advances in Neural
Information Processing Systems}, 2023.

[38] Yujia Qin, Shihao Liang, Yining Ye, Kunlun Zhu, Lan Yan,
Yaxi Lu, Yankai Lin, Xin Cong, Xiangru Tang, Bill Qian, et al.
Toolllm: Facilitating large language models to master 16000+
real-world apis. In \textit{International Conference on Learning
Representations (ICLR)}, 2024.

[39] Guanzhi Wang, Yuqi Xie, Yunfan Jiang, Ajay Mandlekar,
Chaowei Xiao, Yuke Zhu, Linxi Fan, and Anima Anandkumar. Voyager:
An open-ended embodied agent with large language models.
\textit{Transactions on Machine Learning Research}, 2024.

[40] Xuezhi Wang, Jason Wei, Dale Schuurmans, Quoc Le, Ed Chi,
Sharan Narang, Aakanksha Chowdhery, and Denny Zhou.
Self-consistency improves chain of thought reasoning in language
models. In \textit{International Conference on Learning
Representations (ICLR)}, 2023.

[41] Jason Wei, Xuezhi Wang, Dale Schuurmans, Maarten Bosma,
Brian Ichter, Fei Xia, Ed Chi, Quoc V. Le, and Denny Zhou.
Chain-of-thought prompting elicits reasoning in large language
models. In \textit{Advances in Neural Information Processing
Systems}, 2022.

[42] Xuezhi Wang, Jason Wei, Dale Schuurmans, Quoc Le, Ed Chi,
and Denny Zhou. Rationale-augmented ensembles in language
models. \textit{arXiv preprint arXiv:2207.00747}, 2022.

[43] Hunter Lightman, Vineet Kosaraju, Yura Burda, Harrison
Edwards, Bowen Baker, Teddy Lee, Jan Leike, John Schulman, Ilya
Sutskever, and Karl Cobbe. Let's verify step by step. In
\textit{International Conference on Learning Representations
(ICLR)}, 2024.

[44] Karl Cobbe, Vineet Kosaraju, Mohammad Bavarian, Mark Chen,
Heewoo Jun, Lukasz Kaiser, Matthias Plappert, Jerry Tworek, Jacob
Hilton, Reiichiro Nakano, et al. Training verifiers to solve math
word problems. \textit{arXiv preprint arXiv:2110.14168}, 2021.

[45] Mark Chen, Jerry Tworek, Heewoo Jun, Qiming Yuan, Henrique
Ponde de Oliveira Pinto, Jared Kaplan, Harri Edwards, Yuri Burda,
Nicholas Joseph, Greg Brockman, et al. Evaluating large language
models trained on code.
\textit{arXiv preprint arXiv:2107.03374}, 2021.

[46] Eric Zelikman, Yuhuai Wu, Jesse Mu, and Noah D. Goodman.
Star: Bootstrapping reasoning with reasoning. In \textit{Advances
in Neural Information Processing Systems}, 2022.

[47] Caglar Gulcehre, Tom Le Paine, Srivatsan Srinivasan, Ksenia
Konyushkova, Lotte Weerts, Abhishek Sharma, Aditya Siddhant,
Alex Ahern, Miaosen Wang, Chenjie Gu, et al. Reinforced
self-training (rest) for language modeling.
\textit{arXiv preprint arXiv:2308.08998}, 2023.

[48] Avi Singh, John D. Co-Reyes, Rishabh Agarwal, Ankesh Anand,
Piyush Patil, Xavier Garcia, Peter J. Liu, James Harrison, Jaehoon
Lee, Kelvin Xu, et al. Beyond human data: Scaling self-training
for problem-solving with language models. \textit{Transactions on
Machine Learning Research}, 2024.

[49] Edward J. Hu, Yelong Shen, Phillip Wallis, Zeyuan Allen-Zhu,
Yuanzhi Li, Shean Wang, Lu Wang, and Weizhu Chen. Lora: Low-rank
adaptation of large language models. In \textit{International
Conference on Learning Representations (ICLR)}, 2022.

[50] Tim Dettmers, Artidoro Pagnoni, Ari Holtzman, and Luke
Zettlemoyer. Qlora: Efficient finetuning of quantized llms. In
\textit{Advances in Neural Information Processing Systems},
2023.

[51] Woosuk Kwon, Zhuohan Li, Siyuan Zhuang, Ying Sheng, Lianmin
Zheng, Cody Hao Yu, Joseph E. Gonzalez, Hao Zhang, and Ion
Stoica. Efficient memory management for large language model
serving with pagedattention. In \textit{Proceedings of the 29th
Symposium on Operating Systems Principles (SOSP)}, pages 611--626,
2023.

[52] Tri Dao, Daniel Y. Fu, Stefano Ermon, Atri Rudra, and
Christopher R\'e. Flashattention: Fast and memory-efficient exact
attention with io-awareness. In \textit{Advances in Neural
Information Processing Systems}, 2022.

[53] Tri Dao. Flashattention-2: Faster attention with better
parallelism and work partitioning. In \textit{International
Conference on Learning Representations (ICLR)}, 2024.

[54] Tianle Cai, Yuhong Li, Zhengyang Geng, Hongwu Peng, Jason D.
Lee, Deming Chen, and Tri Dao. Medusa: Simple llm inference
acceleration framework with multiple decoding heads. In
\textit{International Conference on Machine Learning (ICML)},
2024.

[55] Tom B. Brown, Benjamin Mann, Nick Ryder, Melanie Subbiah,
Jared Kaplan, Prafulla Dhariwal, Arvind Neelakantan, Pranav Shyam,
Girish Sastry, Amanda Askell, et al. Language models are few-shot
learners. In \textit{Advances in Neural Information Processing
Systems}, 2020.

[56] Hugo Touvron, Louis Martin, Kevin Stone, Peter Albert,
Amjad Almahairi, Yasmine Babaei, Nikolay Bashlykov, Soumya Batra,
Prajjwal Bhargava, Shruti Bhosale, et al. Llama 2: Open
foundation and fine-tuned chat models.
\textit{arXiv preprint arXiv:2307.09288}, 2023.

[57] Thomas Wolf, Lysandre Debut, Victor Sanh, Julien Chaumond,
Clement Delangue, Anthony Moi, Pierric Cistac, Tim Rault, R\'emi
Louf, Morgan Funtowicz, et al. Transformers: State-of-the-art
natural language processing. In \textit{Proceedings of the 2020
Conference on Empirical Methods in Natural Language Processing:
System Demonstrations}, pages 38--45, 2020.

\endgroup

\appendix
\newpage

\section{Related work}
\label{app:related_extended}

\paragraph{Group-relative policy optimisation and its variants.}
GRPO~[1, 2] replaces the learned value baseline of TRPO~[15] /
PPO~[4] with the empirical mean of $G$ rollouts of the same
prompt; both descend from REINFORCE-style policy
gradients~[10, 16]. Several follow-up variants modify the
advantage computation:
RLOO~[17] uses a leave-one-out estimator, ReMax~[18] uses the
maximum reward as the baseline, Dr.GRPO~[19] removes the
$\sigma_r$ normalisation, and REINFORCE++~[20] adds token-level
normalisation. None of these addresses the zero-variance group
problem directly: when all $G$ rewards are equal, any baseline
based on the group itself yields zero advantage. They change the
gradient magnitude on non-zero-variance groups but leave the
wasted rollout cost on zero-variance groups untouched. The
broader RLHF lineage~[21, 22, 3], including DPO~[23],
Constitutional AI~[24] and RLAIF~[25], shares the policy-gradient
backbone but optimises against learned reward or preference
models rather than environment-grounded rewards.

\paragraph{Filtering convergent groups: prompt-level and
in-rollout.}
DAPO~[8] drops zero-variance groups from the gradient batch after
the rollouts have already finished, replacing them with new
prompts to keep the batch full; exact but post-hoc.
GRESO~[9] predicts which prompts will be zero-variance before any
rollout begins, using cross-epoch reward consistency on
mathematical reasoning benchmarks. Both decide at the prompt
level and are orthogonal to a mid-rollout decision. Curriculum-
learning approaches~[26, 27] also reweight prompts at the prompt
level, by difficulty rather than by predicted variance.
We are not aware of prior work that uses intermediate rollout
state to predict zero-variance groups in GRPO. The closest
single-trajectory analogues are early-exit inference: CALM~[28]
terminates a single generation when its per-token confidence
saturates, and speculative decoding~[29, 30] verifies a fast
draft model against the target. These act on one trajectory; our
setting requires a decision over a group of $G$ trajectories
using their disagreement as the signal.

\paragraph{LLM agents and agent RL benchmarks.}
ALFWorld~[5], WebShop~[6], WebArena~[31], and AgentBench~[7] are
standard testbeds for LLM agent behaviour, with
WebGPT~[32] an early instance of agent RL on browser tasks.
They share a structural property that mathematical reasoning
benchmarks lack: every rollout is a multi-turn trajectory
through an observable state space, so the shape of an
in-progress group can be inspected at any intermediate step.
ReAct~[33] interleaves reasoning with action, Reflexion~[34] and
Self-Refine~[35] add verbal self-critique, Tree-of-Thoughts~[36]
explores multiple reasoning branches, and Toolformer~[37] /
ToolLLM~[38] expand the action space with API calls. Voyager~[39]
embeds an LLM agent in an open-ended environment with skill
acquisition. These contributions design \emph{what} the policy
does inside a rollout; ours is about \emph{when to stop} the
rollout.

\paragraph{Multi-sample inference and training efficiency.}
The $G\times$ multi-sample structure also appears at inference
time: self-consistency~[40] takes a majority vote over $G$
chain-of-thought~[41] rollouts, rationale-augmented
ensembles~[42] aggregate weighted samples, best-of-$N$ sampling
with a verifier or reward model picks the best of $N$
candidates~[43, 44], and pass@$k$~[45] is the standard $G$-sample
capacity-evaluation metric. The same divergence signal applies
to these settings. A complementary line of work reuses
trajectories for self-training rather than discarding them:
STaR~[46], ReST~[47] and \emph{Beyond human data}~[48] iterate
on model-generated solutions filtered by reward, which
\emph{exploits} successful rollouts where our gate \emph{cuts}
saturated ones. Orthogonal lines of work reduce per-token cost
via parameter-efficient fine-tuning (LoRA~[49], QLoRA~[50]) and
inference-system optimisations (PagedAttention~[51],
FlashAttention~[52, 53], Medusa~[54]); our gate reduces the
\emph{number} of steps generated rather than the cost of each
step. Throughout we use Qwen2.5-7B-Instruct~[11] as the base
policy, in line with current open-base-model
practice~[55, 56].

\section{Full mid-rollout signal definitions}
\label{app:signals}

The body (\S\ref{sec:method_signals}) defines $d_K$ and points
here for the other six in-group divergence signals we considered.
Each is a scalar in $[0, 1]$ computed from the $G$ partial
trajectories at step $K$, with $0$ meaning the trajectories are
identical at that point.

\begin{itemize}[leftmargin=*,topsep=0pt,itemsep=2pt]
\item \textbf{prefix\_edit\_distance\_mean} ($d_K$). The body
signal: mean over $\binom{G}{2}$ pairs of the Levenshtein edit
distance between $a_{i, 1:K}$ and $a_{j, 1:K}$, divided by the
longer length.
\item \textbf{action\_bigram\_jaccard\_mean}. For each pair of
trajectories, take the set of consecutive action pairs (action
bigrams) in each trajectory's first $K$ steps, compute the Jaccard
overlap of the two sets, and report $1$ minus that overlap.
Average over all pairs. Like $d_K$ this is $0$ when the
trajectories use the same action transitions and grows as they
diverge.
\item \textbf{unique\_prefix\_ratio}. Number of distinct
trajectories among $a_{i, 1:K}$, divided by $G$. The minimum
$1/G$ means all $G$ trajectories share the same first-$K$
sequence; the maximum $1$ means every trajectory is different.
\textbf{unique\_action\_ratio} is the same quantity computed on
the single action $a_{i, K}$ at step $K$ rather than on the full
prefix.
\item \textbf{action\_entropy}. Shannon entropy of the empirical
distribution of $\{a_{i, K}\}$ across the $G$ trajectories at
step $K$.
\item \textbf{obs\_unique\_ratio}. Same as
\textbf{unique\_action\_ratio} but on the environment observation
$o_{i, K}$ rather than the action; a proxy for whether the
trajectories have reached the same world state.
\item \textbf{termination\_fraction} ($\tau_K$). Fraction of the
$G$ trajectories that have already finished (either succeeded or
failed) by step $K$. $\tau_K\!=\!0$ means no trajectory is done,
$\tau_K\!=\!1$ means all $G$ are done. Note this signal is not
itself in $[0, 1]$ as a divergence measure (it is unimodal in the
group label), and we include it only as a complementary candidate
for the OR-rule extension in App.~\ref{app:or_extension}.
\end{itemize}

The full $7\!\times\!4$ side-by-side comparison
(Spearman $\rho$ and AUROC for each metric at
$K\!\in\!\{5, 10, 15, 20\}$) is in
App.~\ref{app:full_correlation}.

\section{Implementation details}
\label{app:impl}

\subsection{How the gate is enforced inside a rollout}
\label{app:supervisor}

The rollout loop generates the $G$ trajectories of a group one
environment step at a time. After the $K$-th environment step we
pause the loop and compute the in-group divergence $d_K$ on the
partial trajectories produced so far. If $d_K$ falls below the
threshold $d_L$, every trajectory in the group is recorded as
terminated at step $K$ with whatever reward it has accumulated
by then, and the rollout loop exits early for that group. The
remaining $T_{\max} - K$ steps are not generated.
\textbf{The accumulated step-$K$ rewards are kept on the
trajectory records only for logging and visualization} (e.g.\ the
training-reward panel of Fig.~\ref{fig:onpolicy} subtracts cut
groups precisely because their truncated reward of $0$ is not a
real outcome); \textbf{the cut group itself is dropped from the
GRPO advantage and loss computation entirely}
(\S\ref{sec:method_gate}, line `\texttt{if is\_cut: continue}' in
the released code), so these truncated rewards never enter the
gradient. When multiple rollouts run in parallel on different
GPUs, each applies the gate independently and the wall-clock
saving on the full set of groups is the sum of per-group savings.
No model parameters are changed during this rollout phase; the
gate only affects how much of each group's trajectories is
generated.

\subsection{Teacher-forcing GRPO loss in agent setting}
\label{app:teacher_forcing}

After rollout, each trajectory $\boldsymbol\tau$ is a token sequence
$[\text{prompt}, o_1, a_1, o_2, a_2, \ldots, o_T, a_T]$ where $o_t$ is
the environment observation at step $t$ and $a_t$ is the model's
response. Only $a_t$ tokens are produced by $\pi_\theta$; $o_t$ tokens
are environment-injected. Computing the policy-gradient loss requires
$\log p_\theta(a_t \mid \text{prefix})$ at every action-token
position.

We compute this with a single teacher-forcing forward pass: the
entire token sequence is fed into the policy model with a causal
attention mask, which returns logits at every position. The logit
at position $t-1$ predicts
the token at position $t$; we extract log-softmax values at each
\emph{action-token} position only — observation-token positions are
masked out, since the policy did not produce them. The trajectory
log-likelihood is
$\log p_\theta(\boldsymbol\tau)\!=\!\sum_{t \in \mathcal{A}}
\log p_\theta(\tau_t \mid \tau_{<t})$ where $\mathcal{A}$ is the set
of action-token positions, and the policy-gradient loss is
$\mathcal{L} = -A \cdot \log p_\theta(\boldsymbol\tau)$.

The action-token mask is constructed at rollout time and stored
alongside the token IDs. Without it, the loss would also penalise
$\theta$ for ``predicting'' the environment's observations, which is
both incorrect (those tokens were not generated by $\theta$) and
empirically destabilising (large gradient-norm spikes when the
environment emits unexpected text).

\subsection{Reloading the policy between iterations}
\label{app:lora_hotswap}

The on-policy training run in Tier 3 alternates between sampling
rollouts and updating the policy. After each gradient update, the
trained LoRA adapter weights are written to disk and then loaded
back into the inference engine before the next iteration's
rollouts begin. The base model itself stays resident in GPU
memory and is not reloaded; only the small LoRA delta is
swapped. The total round-trip cost is on the order of one second
per iteration at LoRA rank $16$, which is negligible compared to
the time spent generating the iteration's rollouts.

\subsection{Hyperparameters}
\label{app:hparams}

Table~\ref{tab:hparams} lists the hyperparameters used in each
of the three integration tiers. The same gate setting
$(K, d_L) = (10, 0.12)$ is used in all three tiers, chosen on the
offline buffer in \S\ref{sec:gate_ablation} and not retuned for
the online runs. Tier 1 has no training so the LoRA and optimiser
hyperparameters do not apply. Tier 2 uses a smaller LoRA rank
($r = 8$, two target modules) because the buffer is fixed and
the optimisation is short ($20$ steps); Tier 3 uses a larger
LoRA rank ($r = 16$, seven target modules) because the policy
must continue to learn over $60$ on-policy iterations. The
sampling temperature during rollout is $0.7$ in all tiers, which
keeps the $G$ trajectories of a group meaningfully different from
each other; held-out evaluation in Tier 3 uses greedy decoding
(temperature $0$) so each test task is decoded once and the
outcome is deterministic.

\begin{table}[h]
\centering
\caption{Hyperparameters across the three integration tiers.}
\label{tab:hparams}
\small
\begin{tabular}{lccc}
\toprule
& Tier 1 (rollout-only) & Tier 2 (off-policy) & Tier 3 (on-policy) \\
\midrule
prompts                    & $100$ (baseline \& gated, same seed) & $100$ (replay buffer) & $60{\times}10$ resampled \\
gradient steps             & $0$            & $20$               & $60$ \\
groups per step            & ---            & $4$                & $10$ \\
action steps used per traj.\ & --- & $8$ (sampled) & all \\
LoRA rank / $\alpha$       & ---            & $8$ / $16$         & $16$ / $32$ \\
LoRA targets               & ---            & $q,v$              & $q,k,v,o,$ MLP \\
optimizer                  & ---            & AdamW              & AdamW \\
learning rate              & ---            & $5{\times}10^{-5}$ & $5{\times}10^{-5}$ \\
sample temp.\ (rollout)    & $0.7$          & $0.7$              & $0.7$ \\
sample temp.\ (eval)       & ---            & ---                & $0$ (greedy) \\
$T_{\max}$                 & $30$           & $30$               & $30$ \\
$G$                        & $8$            & $8$                & $8$ \\
gate $(K, d_L)$            & $(10, 0.12)$ & $(10, 0.12)$ & $(10, 0.12)$ \\
random seed                & $42$           & $42$               & $\{7, 13, 23, 42\}$ \\
\bottomrule
\end{tabular}
\end{table}

\subsection{Setting the precision floor}
\label{app:precfloor}

The $0.80$ precision floor is set so that the worst-case impact on
the un-gated GRPO advantage $L^2$-norm is bounded by a tolerable
fraction $\eta$. Concretely, if the gate's precision is $p$, then the
fraction of \emph{useful} groups (non-zero-variance) it incorrectly
cuts is $(1\!-\!p) \cdot |\text{cut}| / |\text{nonzero-var}|$. For
the main single-axis gate at $d_L\!=\!0.12$:
$|\text{cut}|\!=\!21$, $|\text{nonzero-var}|\!=\!61$,
$p\!=\!0.81$ → fraction lost $\approx 21 \cdot 0.19 / 61 \approx 6.5\%$
of useful groups. The empirical gradient-$L^2$ loss
($3.3\%$, \S\ref{sec:gate_ablation}) is even smaller because the
falsely-cut groups had below-average $|A_i|$. To target a worst-case
$\eta\!=\!10\%$ gradient-norm loss with our group composition, the
precision floor is $1 - \eta \cdot |\text{nonzero-var}| / |\text{cut}|
\!\approx\! 1 - 0.10 \cdot 61/21 \approx 0.71$, well below $0.80$;
the floor is thus a conservative choice.

\subsection{Hardware and timing}
\label{app:hardware}

All experiments use NVIDIA RTX 6000 Ada (48 GB) GPUs. Tier 1
runs the $100$ tasks across $4$ GPUs in parallel. Tiers 2 and 3
each use two GPUs, one dedicated to sampling rollouts and one
dedicated to running the gradient update. Each on-policy
training run (baseline or gated) takes $4.5$ to $4.8$ hours.
Total compute used for the paper is approximately $30$ GPU-hours.
Inference uses vLLM~[51] and gradient updates use the Hugging Face
Transformers~[57] and PEFT/LoRA~[49] stacks.

\section{Extended discussion}
\label{app:discussion_extended}

\subsection{Mechanistic explanation of the held-out gain}
\label{app:dilution_extended}

The held-out improvement we report in Tier 3
(\S\ref{sec:onpolicy}) has two candidate explanations. The first
is that the gate changes the \emph{set} of training prompts in a
way that biases generalisation. The second is that the gate
changes the \emph{magnitude} of each gradient step without
changing its expectation. We have direct evidence for the second
and against the first.

\paragraph{Evidence for the gradient-magnitude explanation.}
The dilution analysis in \S\ref{sec:dilution} predicts a gradient
amplification factor of about $1.20$ at the cut rate we observe.
The measured factor is $1.16 \pm 0.07$ across the $4$ seeds
(Table~\ref{tab:onpolicy}), close to the prediction. The
mechanism is mechanical and predictable: removing zero-advantage
trajectories from the batch shrinks the denominator of the loss
mean while leaving the numerator unchanged.

\paragraph{Evidence against the prompt-distribution explanation.}
When the gate cuts a group, that group is dropped from the
gradient batch but the prompt is not re-sampled. The next training
iteration draws a fresh batch of $10$ prompts uniformly at random
from the same training pool. The gate therefore does not bias the
prompt distribution, and any distribution-shift explanation of the
held-out gain is ruled out by construction.

\paragraph{Practical implication.}
A practitioner who does not want to use the gate could approximate
the held-out gain by raising the learning rate by about $1.16\times$
in the baseline. This approximation is crude, however, because the
zero-advantage fraction varies between iterations (we see $25\%$ to
$70\%$ across the $60$ iterations of our run). The gate scales the
effective learning rate dynamically per iteration, which a fixed
learning-rate increase cannot match.

\subsection{Future directions}

\paragraph{Environment-grounded progress signals for high-$d_K$
all-failure groups.}
The gate catches the low-$d_K$ end of the all-fail cluster
($10/25$ in our buffer; \S\ref{sec:two_axis}) but cannot reach
the high-$d_K$ majority because their early-action prefixes
overlap with mixed-outcome groups in the same $d_K$ region. Detecting them would require a
signal beyond what the partial trajectories themselves provide:
for example, an intermediate progress proxy from the environment
(``object picked up'', ``sub-goal reached'') that distinguishes a
stuck rollout from one that is still exploring. ALFWorld and most
agent benchmarks expose such proxies, but using them is outside
the scope of this paper.

\paragraph{Per-group choice of the gate evaluation step.}
We use a fixed $K = 10$ throughout. Some groups become
unambiguously zero-variance much earlier (the trajectories collide
on identical prefixes by $K = 5$); others are not separable until
$K = 15$. A model that predicts the optimal $K$ for each group
from a few early steps of the rollout could recover further
compute.

\paragraph{Composition with DAPO and GRESO.}
Our gate is orthogonal to both filters. GRESO can skip a prompt
before any rollout begins, our gate can stop a group's rollout
partway through, and DAPO can drop a group from the gradient
batch after the rollout has finished. The three operate at three
distinct stages of the training pipeline and compose naturally.
A natural follow-up is to measure the combined saving end-to-end.

\section{Detailed experimental results}
\label{app:results_extended}

\subsection{Calibration against random-cut and oracle gates}
\label{app:calibration}

We compare our gate against two reference policies that are not
prior work, but do bracket what any cut policy operating at the
same cut rate can do.

The first is a \textbf{random-cut} policy that fires on $20\%$ of
groups uniformly at random, matching our gate's overall cut rate.
This policy destroys $12.4\%$ of the GRPO advantage signal
(measured as the $L^2$-norm of the advantage vector before vs
after cutting), because cutting a non-zero-variance group throws
away real gradient information. Our gate, by contrast, preserves
$96.7\%$ of the advantage signal at the same cut rate, confirming
that it is selecting groups in an information-aware way rather
than dropping rollouts blindly.

The second is an \textbf{oracle} policy that has access to the
ground-truth zero-variance label and cuts every zero-variance
group at step $K\!=\!10$. The oracle sets the upper bound on what
any gate of this form can save: $23.1\%$ of total rollout
step-tokens (raw, using actual rollout step counts). Our gate
recovers $11.3\%$ on the same formula, $\sim\!49\%$ of the
oracle's ceiling. The remaining gap to the oracle is the
high-$d_K$ tail of the zero-variance groups (mostly all-fail):
their early-action prefixes look the same as those of
mixed-outcome groups, so no threshold on prefix divergence
alone can separate them. Closing
this gap would require a different signal, not a better
threshold.

\subsection{Bootstrap confidence intervals on the headline savings}
\label{app:bootstrap_ci}

The Tier~1 A/B run (\S\ref{sec:wallclock}, Table~\ref{tab:wallclock})
is the direct measurement of the gate's wall-clock effect: the
gated arm finishes $13.25\%$ faster than the matched baseline on
the same $100$ ALFWorld tasks. To check that this saving is
robust to the particular set of $100$ tasks we drew, we resample
the $100$ task-level wall-clock differences with replacement
$1{,}000$ times and recompute the percentage saving each time.
The bootstrap $95\%$ confidence interval is $[7.55, 19.12]\%$;
the interval excludes zero.

The offline analysis of \S\ref{sec:gate_ablation} makes a
parallel prediction. Counting step-tokens directly in the
original $100$-task buffer, the gate at $d_L\!=\!0.12$ would have
recovered $11.3\%$ of total step-tokens ($17$ true-positive
groups times the $T_{\max}\!-\!K\!=\!20$ post-$K$ steps each,
divided by the total $100\!\times\!30$ step budget).
Bootstrapping that step-token saving over the same task
resampling gives a $95\%$ confidence interval of $[8.0, 18.0]\%$.
The Tier~1 wall-clock measurement falls inside this predicted
band, confirming that the offline step-token saving translates
into actual GPU wall-clock once the gate is enforced online.

\subsection{Full correlation table}
\label{app:full_correlation}

\begin{table}[h]
\centering
\caption{Spearman $\rho$ of each (metric, $K$) cell against
group reward variance $\sigma_r^2$, $N\!=\!100$, $G\!=\!8$. Cells in
\textbf{bold} exceed the pre-registered threshold
$|\rho|\!\ge\!0.40$. Termination fraction has near-zero monotone
correlation by design (U-shape against label).}
\label{tab:full_corr}
\small
\begin{tabular}{lcccc}
\toprule
metric & $K\!=\!5$ & $K\!=\!10$ & $K\!=\!15$ & $K\!=\!20$ \\
\midrule
prefix\_edit\_distance\_mean ($d_K$) & $0.246$ & $0.374$ & $\mathbf{0.419}$ & $\mathbf{0.418}$ \\
action\_bigram\_jaccard\_mean        & $0.293$ & $\mathbf{0.407}$ & $\mathbf{0.406}$ & $0.389$ \\
unique\_prefix\_ratio                & $0.320$ & $0.398$ & $0.307$ & $0.208$ \\
unique\_action\_ratio                & $0.314$ & $0.379$ & $0.272$ & $0.167$ \\
action\_entropy                      & $0.292$ & $0.369$ & $0.255$ & $0.161$ \\
obs\_unique\_ratio                   & $0.251$ & $0.397$ & $0.307$ & $0.093$ \\
termination\_fraction ($\tau_K$)     & $-0.188$ & $-0.059$ & $0.110$ & $0.235$ \\
\bottomrule
\end{tabular}
\end{table}

\subsection{False-positive case analysis}
\label{app:fp_case}

At $d_L\!=\!0.12$ the gate flags $4$ groups whose final rewards
turn out not to be zero-variance: the gate predicted
convergence but the group ended mixed. All $4$ are
\texttt{pick\_and\_place\_simple} tasks in which the model emits
the same fixed drawer-by-drawer search prefix in every trajectory
of the group, so the gate sees low divergence at $K\!=\!10$ and
fires. The trajectories then diverge \emph{after} step $K\!=\!10$
depending on whether the queried object happens to be in one of
the first few drawers searched: some trajectories find it and
succeed, others do not, so the group's final rewards are mixed
even though its first $10$ actions agree. This points to a
specific failure mode of an action-prefix divergence signal: the
signal cannot detect future divergence in the
\emph{environment-state} branching after the prefix has been
fixed.

\subsection{Per-task-type correlation breakdown}
\label{app:per_type}

Table~\ref{tab:per_type} reports the gate signal for each of the
six ALFWorld task types, in two forms: the global metric
$d_{10}$ that the gate actually uses, and the post-hoc best
$(metric, K)$ pair per type. The global $d_{10}$ generalises
unevenly: it is excellent on
\texttt{look\_at\_obj\_in\_light}, \texttt{pick\_heat}, and
\texttt{pick\_two\_obj} (AUROC $\ge\!0.91$) but degrades on
\texttt{pick\_and\_place\_simple}, \texttt{pick\_clean}, and
\texttt{pick\_cool} (AUROC $0.48$--$0.62$). Allowing the
$(metric, K)$ pair to be tuned per type recovers the worst three
to AUROC $0.71$--$0.86$, with \texttt{termination\_fraction} at
$K\!=\!15$--$20$ being the best signal in the harder pick-cool /
pick-clean tasks. The two task types where the per-type best is
AUROC $1.00$ have only $n_{zv}\!=\!1$--$2$ positives, so those
cells are essentially single-observation; they should be read as
``the gate did not fail'' rather than ``perfect''.

\begin{table}[h]
\centering
\caption{Per-task-type AUROC for the (binary) zero-variance vs
non-zero-variance classification at the published gate's
evaluation step $K\!=\!10$. \textbf{Global $d_{10}$ AUROC} is the
metric the gate actually uses; \textbf{Best AUROC} is the maximum
over $(metric, K)$ tuned per task type, with the winning pair in
parentheses. $n$ = groups in that type;
$n_{zv}$ = zero-variance groups (the positive class).}
\label{tab:per_type}
\small
\begin{tabular}{lccccc}
\toprule
Task type & $n$ & $n_{zv}$ & Global $d_{10}$ AUROC & Best $(metric, K)$ & Best AUROC \\
\midrule
\texttt{pick\_and\_place\_simple}  & $24$ & $13$ & $0.62$ & (unique\_action\_ratio, 15)    & $0.86$ \\
\texttt{pick\_two\_obj\_and\_place}& $20$ & $13$ & $0.91$ & (termination\_fraction, 20)    & $0.86$ \\
\texttt{pick\_cool\_then\_place}   & $19$ &  $5$ & $0.48$ & (termination\_fraction, 20)    & $0.82$ \\
\texttt{pick\_clean\_then\_place}  & $18$ &  $5$ & $0.56$ & (termination\_fraction, 15)    & $0.71$ \\
\texttt{pick\_heat\_then\_place}   & $11$ &  $1$ & $1.00^{*}$ & (action\_entropy, 10)      & $1.00^{*}$ \\
\texttt{look\_at\_obj\_in\_light}  &  $8$ &  $2$ & $1.00^{*}$ & (unique\_prefix\_ratio, 15) & $1.00^{*}$ \\
\midrule
median                              & --- & --- & $0.76$ & ---                           & $0.86$ \\
\bottomrule
\end{tabular}\\
\vspace{2pt}
\footnotesize $^{*}$ $n_{zv}\!\le\!2$, single-observation regime.
\end{table}

\subsection{$G\!=\!8$ vs $G\!=\!16$ robustness panel}
\label{app:g16}

The whole paper uses a group size of $G\!=\!8$. To check that our
findings still hold at a larger group size, we re-collected the
$N\!=\!100$ ALFWorld rollouts with $G\!=\!16$ trajectories per
prompt and recomputed the divergence measures and correlations.

\begin{figure}[h]
\centering
\includegraphics[width=0.7\linewidth]{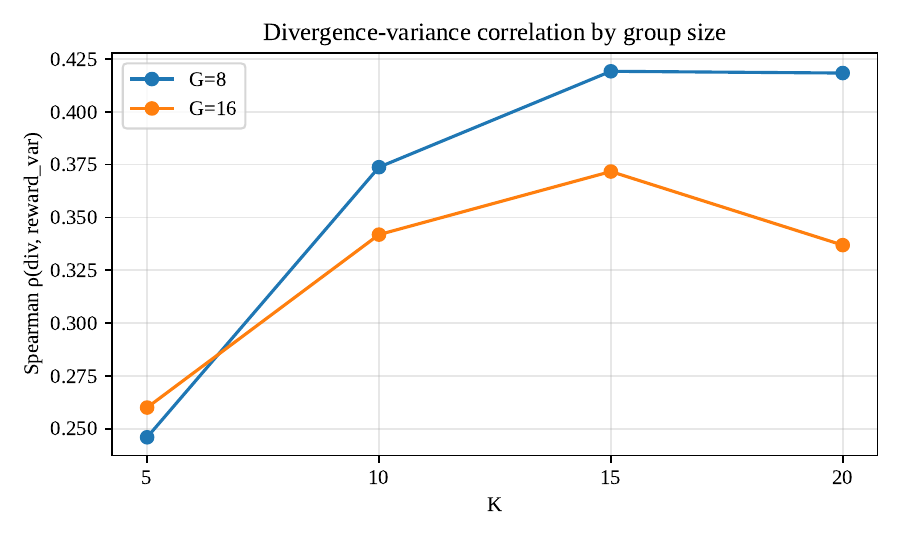}
\caption{Spearman $\rho$ between mid-rollout divergence $d_K$
and final reward variance $\sigma_r^2$, plotted as a function of
the gate evaluation step $K$, for two group sizes ($G\!=\!8$ and
$G\!=\!16$) on the same $100$ ALFWorld tasks. The shape of the
curve is the same in both group sizes; the peak value is slightly
lower at $G\!=\!16$ ($0.37$ vs $0.42$), which we explain in the
text.}
\label{fig:g16}
\end{figure}

Two things stay essentially unchanged. The group-label
composition is similar: $25$ all-fail, $61$ mixed, $14$ all-succeed
at $G\!=\!8$ vs $25$ all-fail, $62$ mixed, $13$ all-succeed at
$G\!=\!16$ ($39$ vs $38$ zero-variance groups out of $100$). The
shape of the correlation curve in $K$ is also similar
(Figure~\ref{fig:g16}): the divergence-variance correlation rises
from $K\!=\!5$, peaks around $K\!=\!10$--$15$, and is roughly flat
or slightly decreasing afterwards, in both group sizes.

The peak correlation is slightly lower at $G\!=\!16$ ($\rho\!=\!0.37$
at $K\!=\!15$) than at $G\!=\!8$ ($\rho\!=\!0.42$). The reason is
mechanical: at $G\!=\!8$ a converged group (whether heading to a
common success or stuck in a common failure) often has all eight
trajectories choose the same first $K$ actions, giving
$d_K\!\approx\!0$ and a clear separation from non-zero-variance
groups; at $G\!=\!16$ it is more common that at least one of the
sixteen trajectories takes a slightly different first action, so
$d_K$ is no longer near zero even when the group ends up
zero-variance. This thins the low-$d_K$ zero-variance cluster
(both all-succeed and all-fail) a little and reduces the rank
correlation.

\paragraph{Threshold transfer is partial.}
Applying the single-axis gate $(K\!=\!10, d_L\!=\!0.12)$ verbatim
to the $G\!=\!16$ data flags more groups but at lower precision,
because with more samples per group it is more likely that a
non-zero-variance group happens to share a low-divergence prefix
by chance. Re-fitting the threshold on $G\!=\!16$ at the same
$0.80$ precision floor moves the operating point to
$(K\!=\!15, d_L\!=\!0.09)$, with savings of about $7.0\%$,
strictly lower than at $G\!=\!8$ because each cut group at
$K\!=\!15$ recovers only $T_{\max}\!-\!K\!=\!15$ remaining steps
instead of $20$. The gate transfers across group sizes, but its
threshold should be retuned at each $G$.

\subsection{Off-policy training per-step trace}
\label{app:offpolicy_trace}

In Tier~2 each training step samples $4$ groups from the buffer.
For every trajectory in those groups (at most $4 \times G = 32$
trajectories per step), the trainer samples up to $8$ action
positions and computes the policy-gradient loss at each one. A
``training item'' in the figure below is one such (trajectory,
action position) pair, so the maximum number of items per step
is $4 \times 8 \times 8 = 128$. In the gated arm a cut group
contributes zero items, so each cut removes up to $8 \times 8 = 64$
items.

\begin{figure}[h]
\centering
\includegraphics[width=0.95\linewidth]{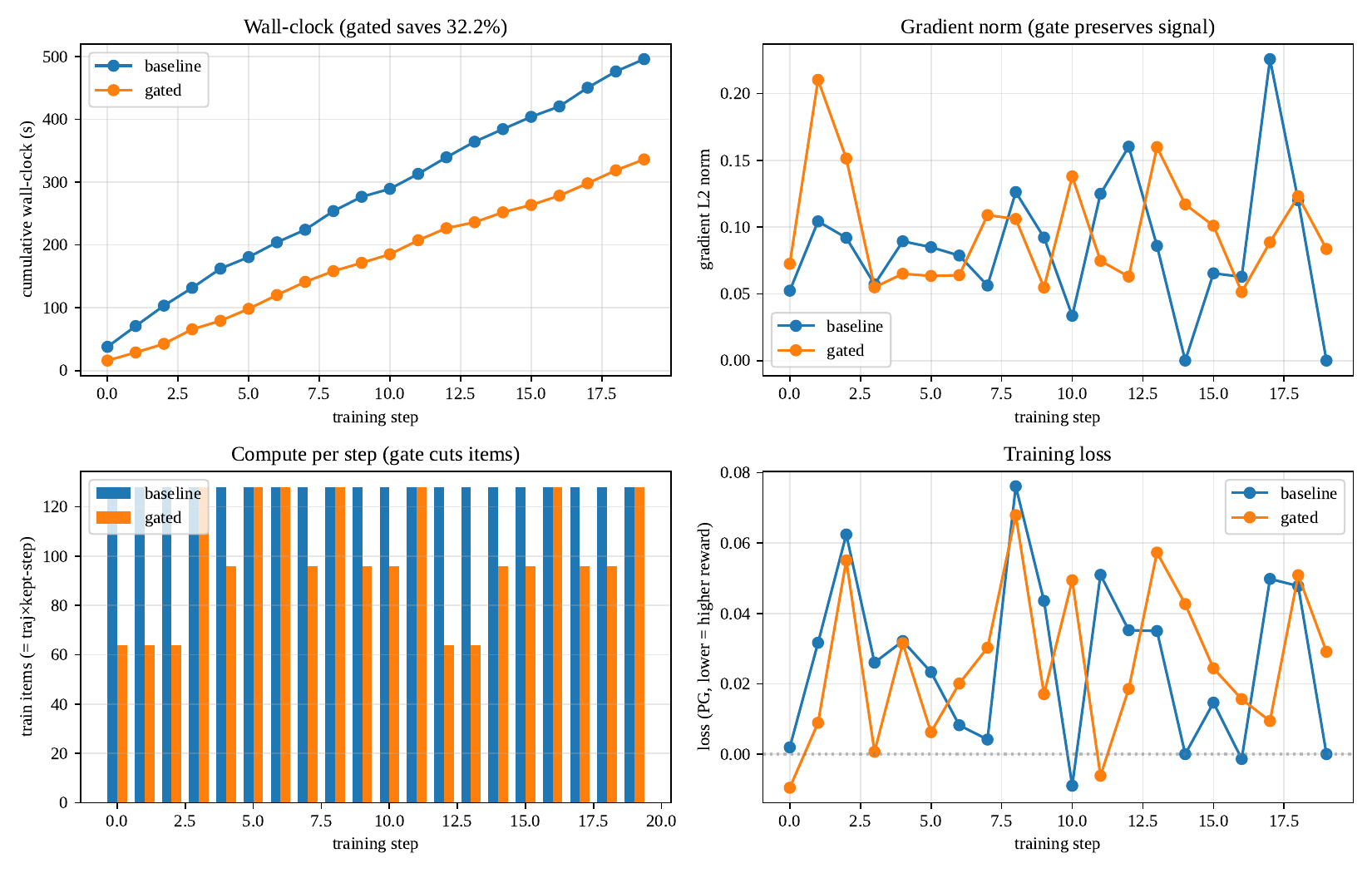}
\caption{$20$-step off-policy GRPO training A/B on Qwen2.5-7B
$+$ LoRA.
\emph{Top left:} cumulative wall-clock; the gated arm finishes
$32.2\%$ faster end-to-end.
\emph{Top right:} per-step gradient $L^2$-norm; the gated arm
runs at about $114\%$ of baseline.
\emph{Bottom left:} number of training items per step (one item
$=$ one trajectory $\times$ one action position; max $128$ as
explained above). The baseline arm always uses the maximum $128$
items because no group is cut. The gated arm drops to between
$64$ and $128$ items per step depending on how many groups the
gate cut at that step.
\emph{Bottom right:} per-step training loss. Both arms sit in the
same noisy band around zero. This is expected: by definition the
GRPO advantages within a group sum to zero, so the population
loss for the population gradient estimator is zero up to noise.
A loss near zero in this plot does not mean the policy stopped
learning; held-out evaluation in Tier~3 is the metric that
captures policy quality.}
\label{fig:training_trace}
\end{figure}

\subsection{On-policy training per-iter trace and held-out trajectory}
\label{app:onpolicy_trace}

In Tier~3, every $10$ iterations we pause training and run greedy
decoding on $50$ held-out ALFWorld tasks (\S\ref{sec:onpolicy}).
Table~\ref{tab:onpolicy_eval} shows the resulting success rate at
each of the $7$ evaluation points (iters $0, 10, 20, \dots, 60$),
averaged over the $4$ random seeds we ran. At iter $0$ no training
has happened yet, so the gated and baseline arms share the same
policy by construction and therefore evaluate identically per seed
(the across-seed std is non-zero because the held-out task set is
seeded from the run's random seed). The held-out gap between the
two arms is non-monotonic across iterations: gated is slightly
behind at iters $20$--$30$ and slightly ahead from iter $40$
onward. The $+2.5$~pp mean gap we report at iter $60$ in the main
text is one point on this noisy trajectory.

\begin{table}[h]
\centering
\caption{Held-out success rate across the $60$ on-policy
iterations, averaged over the $4$ seeds. $\Delta$ is the
per-seed mean of (gated $-$ baseline). At iter $0$, gated and
baseline are the same policy (same random seed), so per-seed
they coincide and $\Delta = 0$ exactly. Across-seed standard
deviations are reported in Figure~\ref{fig:onpolicy} (shaded
bands) and Table~\ref{tab:onpolicy} (iter $0$ and iter $60$).}
\label{tab:onpolicy_eval}
\small
\begin{tabular}{lccccccc}
\toprule
iter      & 0 & 10 & 20 & 30 & 40 & 50 & 60 \\
\midrule
baseline  & $39.5\%$ & $38.5\%$ & $39.5\%$ & $40.5\%$ & $39.0\%$ & $37.0\%$ & $41.5\%$ \\
gated     & $39.5\%$ & $39.5\%$ & $38.0\%$ & $37.5\%$ & $43.5\%$ & $40.0\%$ & $44.0\%$ \\
$\Delta$  & $0$      & $+1.0$   & $-1.5$   & $-3.0$   & $+4.5$   & $+3.0$   & $+2.5$   \\
\bottomrule
\end{tabular}
\end{table}

\paragraph{Per-iter gate firing pattern.}
The per-iter cut count is highly variable, ranging from $0$ to
$6$ groups (out of the $10$ in each iteration) depending on the
difficulty of the iteration's sampled prompts. Across the $4$
seeds and $60$ iterations the gate fires on a mean of
$\mathbf{106\!\pm\!18}$ groups out of $600$ total ($17.7\%$ cut
rate; per-seed totals
$88, 94, 119, 124$ for seeds $13, 7, 23, 42$). The mean baseline
zero-variance rate over the same iters is $40.6\%$, and the
gate's per-iter recall on the on-policy zero-variance ground
truth averages $\sim\!37\%$, comparable to the
$17/39\!\approx\!44\%$ recall on the offline buffer
(Table~\ref{tab:ablation}), indicating the gate's selectivity is
preserved under the on-policy distribution shift.

\paragraph{Has the policy converged?}
The training reward variance does not collapse over the $60$
iterations in either arm, so neither policy has reached a fixed
point within the compute budget we used. The wall-clock saving is
a per-iteration percentage and should therefore stay roughly
constant ($\sim\!10.7\%$) regardless of how many iterations we
run. The held-out gap between the gated and baseline arms,
however, could either widen with more iterations (if the
gradient-amplification effect from \S\ref{sec:dilution} keeps
the gated arm slightly ahead at every step) or shrink to zero
(if both arms eventually reach the same converged policy). The
$n\!=\!4$ seeds and $60$ iterations we run cannot distinguish
these two scenarios; longer training runs are left to future
work.

\subsection{The OR-rule extension: optional termination-fraction backup}
\label{app:or_extension}

The main paper uses a one-parameter single-axis gate
$\textsc{Cut}_{K, d_L}\!:$ \texttt{cut iff} $d_K\!<\!d_L$
(\S\ref{sec:method_gate}). We document here a two-parameter
\emph{OR-rule} extension we explored:
\begin{equation}
\textsc{Cut}^{\lor}_{K, d_L, \tau_H}: \;\;
d_K\!<\!d_L \;\lor\; \tau_K\!\ge\!\tau_H,
\label{eq:or_rule}
\end{equation}
which adds a backup clause that triggers when most of the group
has already terminated. The motivation is the U-shape between
$\tau_K$ and the group label (Table~\ref{tab:ushape_app} below): all-success
groups concentrate at high $\tau_K$ \emph{and} low $d_K$; high $\tau_K$
contains a few extra all-success groups whose trajectories arrived at
the goal via slightly different short prefixes that the
$d$-threshold misses.

\begin{table}[h]
\centering
\caption{Median of $d_K$ and $\tau_K$ at $K\!=\!15$ stratified by
group label, $G\!=\!8$. The two zero-variance classes (all-fail,
all-succeed) sit on \emph{opposite} ends of $\tau_K$, motivating the
U-shape interpretation that originally led us to the OR-rule.}
\label{tab:ushape_app}
\begin{tabular}{lcccc}
\toprule
group label & $n$ & median $d_{15}$ & median $\tau_{15}$ & physical meaning \\
\midrule
all\_succeed & 14 & 0.19 & 1.00 & solved fast, all terminated \\
mixed        & 61 & 0.47 & 0.13 & exploring different strategies \\
all\_fail    & 25 & 0.26 & 0.00 & nobody finishes, often lock-stepped \\
\bottomrule
\end{tabular}
\end{table}

\paragraph{Quantitative effect of adding the $\tau$ backup.}
At $K\!=\!10$, $\tau_H\!=\!0.90$, $d_L\!=\!0.12$, the OR-rule
extension flags $23$ groups (vs $21$ for $d_K\!<\!0.12$ alone) and
hits all $2$ extra groups as true positives, yielding precision
$0.83$ vs $0.81$ and recall $0.49$ vs $0.44$.  \textbf{However, the
incremental rollout savings is exactly zero}: by step $K\!=\!10$
those $2$ extra groups have $\tau\!\ge\!0.90$ — i.e., $7$--$8$ of
their $G\!=\!8$ trajectories have \emph{already} terminated, so
``cutting'' them at $K$ saves nothing the early-termination
mechanism didn't already save (Table~\ref{tab:strong_baselines_app}).

We therefore recommend the single-axis gate as the production
default: simpler, less prone to overfitting on small validation
sets ($N\!=\!100$ here), and indistinguishable in rollout savings.
The OR-rule is documented for completeness and would be the natural
extension on environments where $\tau_K$ at moderate $K$ is more
discriminative than ALFWorld provides.

\paragraph{Trying to catch all-failure groups.}
The OR-rule above adds a backup clause for the all-success end of
the U-shape ($\tau_K \!\ge\! 0.90$). The natural mirror, which
would catch all-failure groups, is to add a backup clause for the
low-$\tau$ end:
$d_K\!<\!d_L \;\lor\; \tau_K\!\le\!t_L$. We swept this rule
across $d_L \in [0.02, 0.30]$, $t_L \in \{0, 0.05, 0.10, 0.15\}$,
and $K \in \{5, 10, 15, 20\}$. \emph{No operating point clears
the $0.80$ precision floor at any $K$}. The reason is geometric:
in early rollout phases mixed groups also have
$\tau_K \approx 0$ (some trajectories are still running, none
has reached a terminal state yet), so any cut on low $\tau_K$
pulls many mixed groups in along with the all-failure ones it
was meant to catch. This is the empirical evidence behind the
``no operating point clears 0.80 precision'' claim in the
contributions section of the main paper.

\subsection{OR-rule training A/B (seed $42$, R3)}
\label{app:or_training}

Before pivoting to the single-axis R1 gate, we ran the on-policy
training A/B with the two-parameter OR-rule extension R3
(App.~\ref{app:or_extension}) on seed $42$.  We report those
numbers here for completeness:

\begin{table}[h]
\centering
\caption{On-policy A/B with the OR-rule extension R3
($d_K\!<\!0.12 \lor \tau_K\!\ge\!0.90$), seed $42$, $60$ iters.
Setup otherwise matches Tier 3 in the main paper.}
\label{tab:or_training}
\small
\begin{tabular}{lrrr}
\toprule
& baseline & gated (R3) & $\Delta$ \\
\midrule
total wall-clock (s)                  & $17{,}252$ & $15{,}364$ & $-10.94\%$ \\
groups cut / total                    & $0/600$  & $96/600$ & cut rate $16.0\%$ \\
zero-advantage items in batch         & $42.0\%$ & $25.7\%$ & $-16.3$ pp \\
gradient $L^2$-norm mean              & $0.123$ & $0.181$ & $+47\%$ \\
\textbf{held-out eval, iter $0/60$}   & $38.0\%/38.0\%$ & $\mathbf{38.0\%/46.0\%}$ & $\mathbf{+8\,\text{pp}}$ \\
bootstrap $95\%$ CI on $\Delta$       &           & $[+2, +16]$ pp & CI excludes $0$ \\
\bottomrule
\end{tabular}
\end{table}

On this single seed (seed 42), the OR-rule extension gives a
$+8$ pp held-out improvement over its baseline. The single-axis
gate on the same seed gives $+2$ pp, and across the four
single-axis seeds we report in the main paper (seeds
$\{7, 13, 23, 42\}$) the held-out improvement is
$+2.5\!\pm\!3.4$ pp on average. The OR-rule's larger improvement
on this single seed is consistent with its larger dilution
reduction: it cuts about $16.0\%$ of groups (vs $17.7\%$ for
single-axis), and its zero-advantage-item rate drops by $-16.3$
pp (vs about $-12$ pp for single-axis), giving a larger gradient
amplification ($1.47\times$ vs $1.16\!\pm\!0.07\times$).

We caution that this OR-rule run is on a single seed and that
the OR-rule has two parameters tuned on the same $N\!=\!100$
offline buffer, so the comparison overstates the OR-rule's true
held-out improvement relative to single-axis. Without OR-rule
runs on additional seeds, we cannot say how much of the
$+8$ pp vs $+2.5$ pp gap is the rule and how much is seed
variance. We report the OR-rule numbers here for completeness
and as an upper-bound reference for what the gate's
policy-improvement effect could look like under more aggressive
cutting; we use the single-axis gate for all main-paper claims.

\subsection{Strong-baseline comparison detail}
\label{app:strong_baselines_app}

\begin{table}[h]
\centering
\caption{Strong-baseline arm comparison on the same $N\!=\!100$
buffer; cut count, TP/FP composition, precision, rollout
step-tokens saved, and GRPO advantage $L^2$-norm preserved
(ratio to no-gate). Random-cut at the same $\sim\!23$-cut budget
destroys $\sim\!12$ pp of $L^2$ because its FPs land on
non-zero-variance groups whose advantages are real. Oracle (cut
iff truly zero-variance) sets the upper bound at $23.1\%$
saving; our single-axis gate $d_K\!<\!0.12$ recovers $51\%$ of
that at $96.7\%$ $L^2$-preservation. DAPO-oracle and
OURS$+$DAPO save the same rollout step-tokens as their
pre-DAPO counterparts because DAPO is a post-rollout filter
that drops groups from the gradient batch only; it changes
the training-phase cost (App.~\ref{app:offpolicy_trace}), not
the rollout-phase cost. The single-axis $\tau_K \ge 0.90$ clause
on its own saves zero rollout compute because the groups it
fires on have already terminated by step $K$.}
\label{tab:strong_baselines_app}
\small
\begin{tabular}{lcccccc}
\toprule
arm & cut & TP & FP & precision & rollout-saved (\%) & $L^2$-preserved (\%) \\
\midrule
no-gate                                 & $0$  & $0$  & $0$  & ---    & $0.0$  & $100.0$ \\
random-cut (matched $23$)               & $23$ & $9$  & $14$ & $0.39$ & $13.3$ & $87.6$ \\
oracle (cut iff zv)                     & $39$ & $39$ & $0$  & $1.00$ & $\mathbf{23.1}$ & $100.0$ \\
DAPO-oracle (post-hoc filter only)      & $0$  & $0$  & $0$  & ---    & $0.0$  & $100.0$ \\
\textbf{OURS R1} ($d_K\!<\!0.12$)        & $21$ & $17$ & $4$  & $0.81$ & $\mathbf{11.3}$ & $\mathbf{96.7}$ \\
single-axis $\tau_K\!\ge\!0.90$ only    & $6$  & $6$  & $0$  & $1.00$ & $0.0$  & $100.0$ \\
OR-rule extension R3 ($d\!\lor\!\tau$)  & $23$ & $19$ & $4$  & $0.83$ & $11.3$ & $96.7$ \\
OURS R1 $+$ DAPO                        & $21$ & $17$ & $4$  & $0.81$ & $11.3$ & $96.7$ \\
\bottomrule
\end{tabular}
\end{table}

DAPO-oracle (post-hoc only) saves $0\%$ rollout compute by
construction — DAPO drops zero-variance groups from the gradient
batch but does not interrupt the rollout. In the off-policy training
A/B (Tier 2, Sec.~\ref{sec:offpolicy}) DAPO-oracle would save the
forward+backward of zero-variance groups, which translates to a
training-phase saving (we measure $32.2\%$ wall-clock there).
\textbf{OURS R1 $+$ DAPO} is the recommended combination for an
end-to-end production trainer: cut early on the divergence signal,
then drop any surviving zero-variance groups from the gradient batch.

\subsection{Mechanism evolution: zv-rate, dilution, and gate recall over training}
\label{app:mechanism_evolution}

\begin{figure}[h]
\centering
\includegraphics[width=0.95\linewidth]{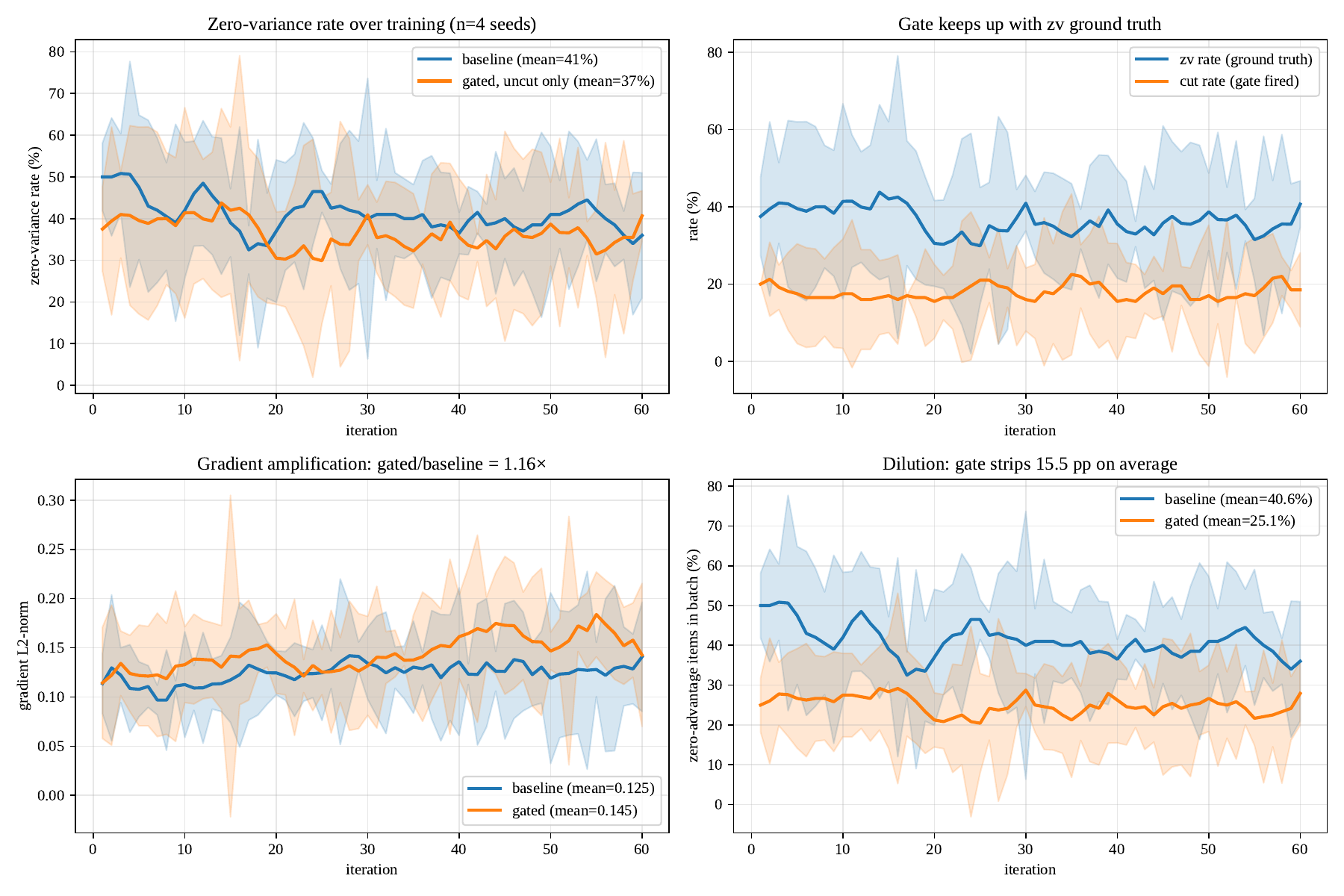}
\caption{Mechanistic evolution across the $60$-iteration on-policy
GRPO training, averaged over $n\!=\!4$ seeds $\{7, 13, 23, 42\}$;
shaded bands are $\pm 1$ std across seeds.
\emph{Top-left:} per-iter zv rate (5-iter rolling) drifts downward
in the baseline as the policy learns; gated tracks closely on the
uncut groups. \emph{Top-right:} gate cut rate (orange) vs
ground-truth zv rate (blue) per iter; per-iter recall averages
$37.2\%$. \emph{Bottom-left:} gradient $L^2$ norm — gated runs
$1.16\!\times$ baseline on average ($1.16\pm 0.07$ across seeds),
matching the dilution prediction
(\S\ref{sec:dilution}, Table~\ref{tab:onpolicy}).
\emph{Bottom-right:} zero-advantage-item rate in the train batch:
baseline $40.6\%$, gated $25.1\%$, a $-15.5$ pp reduction which
is the proximate mechanism for the held-out gain.}
\label{fig:mech_evolution}
\end{figure}

The top-left panel of Figure~\ref{fig:mech_evolution} plots the
per-iteration zero-variance rate in both arms, smoothed with a
$5$-iteration rolling average. The baseline rate fluctuates
between $10\%$ and $90\%$ across iterations and drifts slowly
downward (mean $42\%$), consistent with the policy gradually
moving from all-fail-dominated zero-variance groups (more common
early in training) toward more mixed-outcome groups as it learns
to succeed on at least some of the $G$ rollouts.

For the gated arm we compute the zero-variance rate over the
\emph{uncut} groups only. We exclude cut groups from this
statistic because the gate stops their trajectories at step
$K\!=\!10$ before any reward has arrived in ALFWorld (most
successes happen at step $15$--$25$), so their recorded rewards
are all zero and they would be miscounted as zero-variance even
when their full rollout would have been mixed. With this
correction, the gated zero-variance rate is $37.1\%$ on
average, modestly below baseline as expected: the gate has
removed some genuinely zero-variance groups from the rollout
distribution, leaving a slightly less zero-variance-heavy
remainder.

The top-right panel anchors the gate's recall on the on-policy
distribution. The blue line is the baseline-arm ground-truth
zero-variance rate (the gate had no effect on baseline rollouts);
the orange line is the per-iteration cut rate in the gated arm.
The gray band between them is the recall gap: ground-truth
zero-variance groups that the gate did not cut. On average, the
gate's per-iteration recall against this ground truth is
$34.8\%$, lower than the offline $0.49$ we report in
\S\ref{sec:gate_ablation} because the on-policy rollout
distribution differs from the offline buffer that the threshold
was tuned on. The gate behaves conservatively under this shift,
which is why its precision stays high.

\subsection{Token-level dilution measurement}
\label{app:dilution_appendix}

We estimate the zero-advantage dilution rate directly from the
per-iteration training logs of the on-policy run. For each
iteration we know how many groups were rolled out, how many ended
up zero-variance, and how many were cut by the gate. With
$G = 8$ trajectories per group and $10$ groups per iteration,
the un-gated baseline's zero-advantage fraction averages
about $42\%$, and the gated run's zero-advantage fraction (from
the zero-variance groups that survived the gate) averages about
$26\%$. The $\approx\!16$ percentage-point reduction is what drives
the gradient-norm amplification reported in \S\ref{sec:dilution}.

\section{Prompt templates}
\label{app:prompts}

\paragraph{ALFWorld system prompt.}
We use a ReAct-style prompt format. The system message describes
the ALFWorld environment, lists the canonical action vocabulary,
and provides one in-context example trajectory. The user message
at each step contains the current observation, and the assistant
is expected to emit text in the format
``\texttt{Thought: ... Action: <action>}''. The action string is
parsed greedily; if it does not match the environment's action
grammar, the step is treated as a no-op and still counts towards
$T_{\max}$.

\paragraph{Eval prompt.}
Held-out eval (Tier 3) uses the same prompt format but with sampling
temperature $T\!=\!0$ (greedy) and no gate. A task is counted as
solved iff the environment returns $r\!=\!1$ within $T_{\max}\!=\!30$
steps.

\section{Broader impacts}
\label{app:broader}

The selective-rollout gate reduces the compute footprint of agent RL
training by $11$--$32\%$ on our benchmark, with no degradation in
downstream policy quality — and in the on-policy setting, with a
measurable improvement. The direct effect is to lower the energy and
GPU-time cost of LLM agent training, which aligns with the broader
goal of making such training accessible to research groups without
hyperscale compute. We do not foresee novel dual-use risks beyond
those already inherent to LLM agent RL: any compute-saving technique
applied to capability research generally accelerates capability
research, including capabilities that may have negative externalities
(e.g.\ autonomous web agents executing harmful actions). The gate
itself is agnostic to the reward function and does not introduce new
incentives.

\end{document}